\pdfoutput=1  % Commented out for XeLaTeX compilation
\documentclass[10pt]{article}

\usepackage[preprint]{acl}

\usepackage{times}
\usepackage{latexsym}
\usepackage{amsmath}
\usepackage{amssymb}
\usepackage{xcolor}
\usepackage{booktabs}
\usepackage{microtype}
\usepackage{tcolorbox}
\usepackage{inconsolata}

\usepackage{caption}
\usepackage{graphicx}

\usepackage{hyperref}
\hypersetup{
    colorlinks=true,
    linkcolor=blue,
    citecolor=blue,
    urlcolor=blue
}

\let\oldref\ref
\renewcommand{\ref}[1]{%
    \textcolor{blue}{\oldref{#1}}%
}

\usepackage{geometry}
\usepackage{titlesec}
\titleformat{\section}{\normalfont\normalsize\bfseries}{\thesection}{1em}{}
\titleformat{\subsection}{\normalfont\normalsize\bfseries}{\thesubsection}{1em}{}
\titleformat{\subsubsection}{\normalfont\normalsize\bfseries}{\thesubsubsection}{1em}{}
\titleformat{\paragraph}{\normalfont\normalsize\bfseries}{\theparagraph}{1em}{}

\title{MicroRCA-Agent: Microservice Root Cause Analysis Method Based on Large Language Model Agents}

\author{
    Pan Tang\textsuperscript{\rm 1}, Shixiang Tang\textsuperscript{\rm 1}, Huanqi Pu\textsuperscript{\rm 1}, {Zhiqing Miao\textsuperscript{\rm 2}, Zhixing Wang\textsuperscript{\rm 3}}\\
    \small
    \textsuperscript{\rm 1}School of Communication and Information Engineering, Shanghai University, Shanghai, China\\
    \small
    \textsuperscript{\rm 2}School of Communication and Electronic Engineering, East China Normal University, Shanghai, China\\
    \small
    \textsuperscript{\rm 3}School of Information and Electronics, Beijing Institute of Technology, Beijing, China\\
    \small
    \texttt{ptang@shu.edu.cn}, \texttt{tangsx624@163.com}, \texttt{1287056469@qq.com}, \texttt{52275904009@stu.ecnu.edu.cn}, \texttt{wangwangzx@bit.edu.cn}
}

\renewcommand{\theparagraph}{\alph{paragraph}}

\makeatletter
\renewcommand{\paragraph}{%
    \@startsection{paragraph}{4}{\z@}%
    {3.25ex \@plus1ex \@minus.2ex}%
    {1.5ex \@plus.2ex}%
    {\normalfont\normalsize\itshape}%
}
\makeatother

\begin{document}
\maketitle
\begin{abstract}

This paper presents MicroRCA-Agent, an innovative solution for microservice root cause analysis based on large language model agents, which constructs an intelligent fault root cause localization system with multimodal data fusion\footnote{This work is a technical report of our solution at the 2025 (8th) CCF International AIOps Challenge. The official website of the Challenge is \url{https://challenge.aiops.cn}.}. The technical innovations are embodied in three key aspects: First, we combine the pre-trained Drain log parsing algorithm with multi-level data filtering mechanism to efficiently compress massive logs into high-quality fault features. Second, we employ a dual anomaly detection approach that integrates Isolation Forest unsupervised learning algorithms with status code validation to achieve comprehensive trace anomaly identification. Third, we design a statistical symmetry ratio filtering mechanism coupled with a two-stage LLM analysis strategy to enable full-stack phenomenon summarization across node-service-pod hierarchies. The multimodal root cause analysis module leverages carefully designed cross-modal prompts to deeply integrate multimodal anomaly information, fully exploiting the cross-modal understanding and logical reasoning capabilities of large language models to generate structured analysis results encompassing fault components, root cause descriptions, and reasoning trace. Comprehensive ablation studies validate the complementary value of each modal data and the effectiveness of the system architecture. The proposed solution demonstrates superior performance in complex microservice fault scenarios, achieving a final score of 50.71. The code has been released at: \url{https://github.com/tangpan360/MicroRCA-Agent}.

\end{abstract}

\section{Overall Design}
\label{sec:design}

The proposed solution adopts a modular architectural design, with the overall system comprising five core modules: data preprocessing module, log fault extraction module, trace anomaly detection module, metric fault summarization module, and multimodal root cause analysis module. The modules are designed with loose coupling, facilitating data interaction through function encapsulation, which ensures both system integrity and module independence with enhanced scalability.

The data preprocessing module undertakes the critical responsibility of system data standardization. Its core functionalities include: parsing fault time period information from input.json, and standardizing timestamp formats across log, trace, and metric modalities to ensure accuracy in subsequent cross-modal data correlation.

The log fault extraction module is based on a "template + rule" processing architecture. It first designs multiple regular expression rules covering different fault types. The module utilizes the pre-trained Drain model\cite{he2017drain} (error\_log-drain.pkl) for automatic log template extraction, merging log entries with the same semantics into templates with their occurrence counts through template recognition and deduplication mechanisms, with precise control over template matching parameters via the drain3.ini configuration file. The module implements a multi-level data filtering mechanism: file localization → time window filtering → error keyword filtering → extraction and reconstruction of core fields → fault template matching and standardization → sample deduplication and frequency statistics → name mapping and service extraction, effectively compressing massive raw logs into high-quality fault features.

\begin{figure*}[t]
    \centering
    \includegraphics[width=0.95\linewidth]{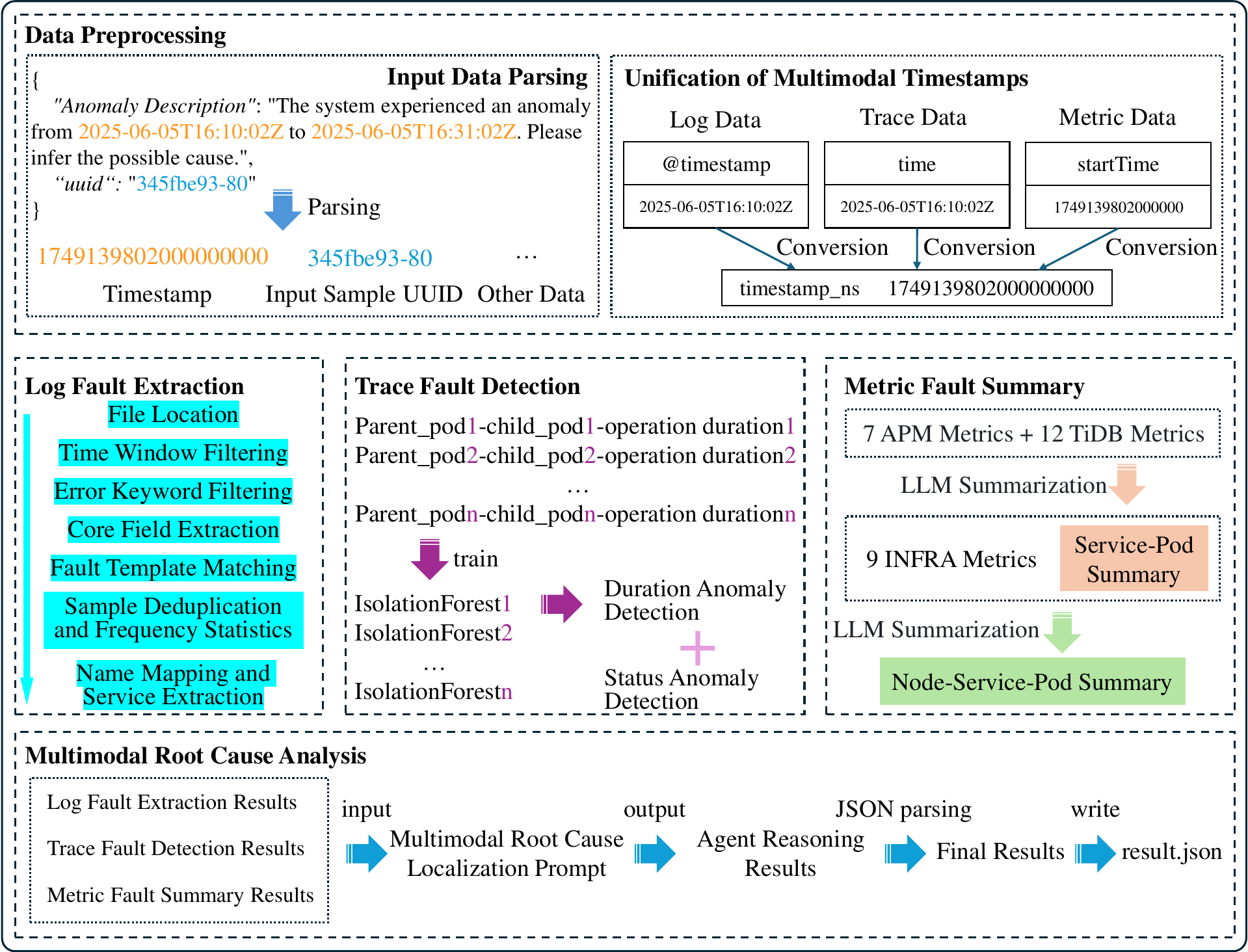}
    \caption{Overall architecture of MicroRCA-Agent.}
    \label{fig1}
\end{figure*}

\begin{figure*}[htbp]
    \centering
    \includegraphics[width=0.95\linewidth]{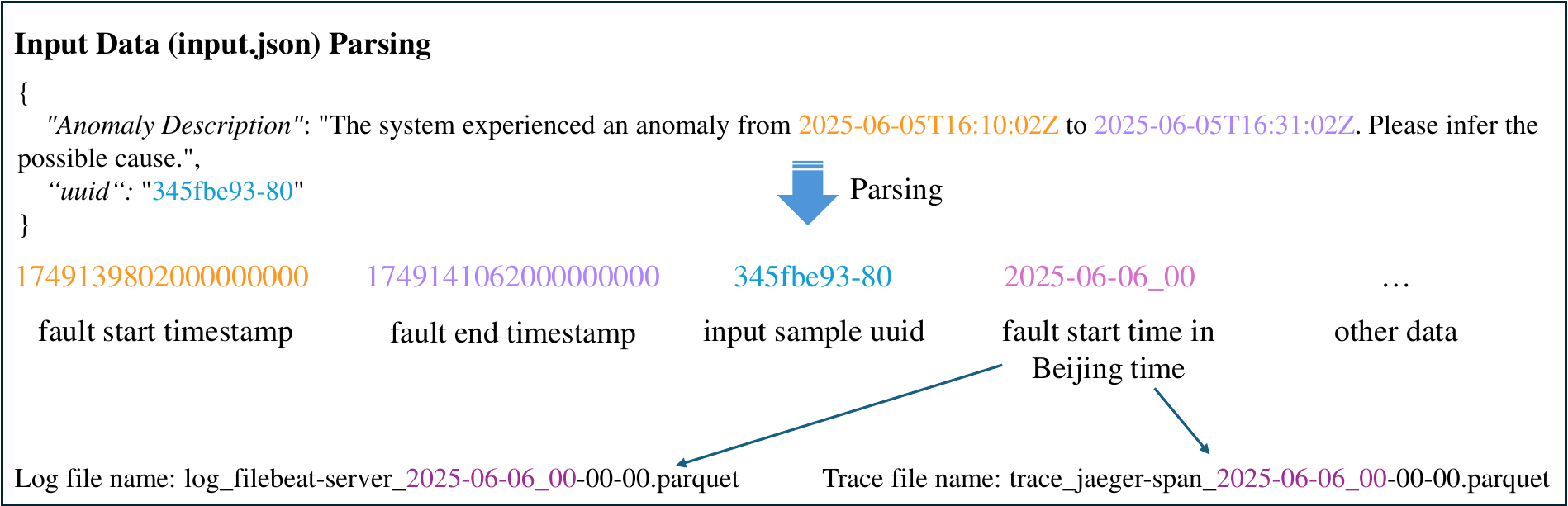}
    \caption{Input data parsing.}
    \label{fig2}
\end{figure*}

\begin{figure*}[htbp]
    \centering
    \includegraphics[width=0.95\textwidth]{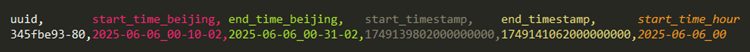}
    \caption{Preprocessed input data.}
    \label{fig3}
\end{figure*}

The trace fault detection module adopts a hybrid detection strategy of "unsupervised learning + rule enhancement". At its core, it leverages the Isolation Forest\cite{liu2008isolation} algorithm to construct anomaly detection model, which uses duration data from parent\_pod→child\_pod call chains during normal periods for model training. The model training process involves extracting normal data from 50 random samples, using 30-second sliding windows to extract temporal features, and generating the trace\_detectors.pkl model file upon completion. During the detection phase, anomaly detection is performed on the duration of parent\_pod→child\_pod call chains in fault periods to filter out significantly anomalous call chains and their corresponding durations. Additionally, it implements a status analysis function, extracting status.code and status.message information from traces to identify call failure patterns. Finally, it outputs two sets of results respectively: statistical analysis of the top 20 most frequent anomalous call combinations, and detailed status anomaly information for the top 20 most frequent cases.

The metric fault summarization module designs a statistical symmetric ratio filtering strategy combined with a hierarchical two-stage LLM analysis approach. First, it performs mathematical statistical analysis on metric indicators and uses symmetric ratios to determine whether the statistical indicators of normal intervals and fault intervals are highly similar, thereby filtering normal data and significantly reducing the large model context length, enabling it to focus more on potentially anomalous data. After filtering, it conducts two-stage LLM analysis: the first stage is based on Application Performance Monitoring (APM) data and database component (TiDB) data, performing phenomenon summarization at both service level and pod level, including business metrics such as microservice request response and anomaly ratios, as well as related indicators of database auxiliary components TiDB, TiKV, and PD. The second stage combines the business metric phenomenon summary from the first stage with node-level infrastructure data to conduct comprehensive phenomenon analysis across three levels: service, pod, and node. This approach effectively controls computational costs while ensuring analytical depth.

The multimodal root cause analysis module is designed with specialized multimodal prompts at its core, supporting flexible combinations of log, trace, and metric data across three modalities. The module adopts a multi-processing strategy to improve system processing efficiency and integrates comprehensive fault tolerance mechanisms and retry strategies. It ultimately outputs structured root cause analysis results containing component, reason, and reasoning trace, achieving a complete closed loop from phenomenon observation to root cause inference.

\section{Innovation and Practicality}
\label{sec:innovation}

The innovation of this solution is embodied in its multi-level feature extraction and reasoning architecture design. First, the log, trace, and metric modules each adopt highly targeted processing and filtering strategies: the Drain algorithm is combined with multi-level data filtering to compress log redundancy; Isolation Forest and status code checking complement each other to identify trace anomalies; statistical significance-based filtering is applied to extract high-value metric features. This modality-specific "precise extraction-information compression" mechanism, tailored to each data type (log, trace, metric), significantly reduces the computational and contextual overhead of subsequent analysis while ensuring the integrity of key information. Second, the metric analysis component incorporates a statistical symmetric ratio filtering mechanism and a two-stage phenomenon summarization framework. After filtering out potentially anomalous data, the framework organically integrates service/instance and container/node perspectives, while preserving the correlation between business operations and infrastructure at the phenomenon level, thereby providing semantically interpretable evidence directly usable for subsequent root cause localization.

In terms of practicality, this solution emphasizes a modular and scalable architectural design, enabling flexible adaptation to different business and system environments. The functional modules (log extraction, trace detection, metric summarization, root cause analysis) are decoupled from one another, allowing for independent deployment or on-demand combination. This design facilitates rapid implementation in microservice architectures of varying scales and complexities. In terms of processing efficiency, the solution significantly reduces the amount of irrelevant data entering the analysis phase via strategies like hierarchical filtering. This reduction in irrelevant data lowers computational and storage overhead, thus making the solution well-suited for high-concurrency, large-data-volume production environments.

\begin{figure*}[ht]
    \centering
    \includegraphics[width=0.8\textwidth]{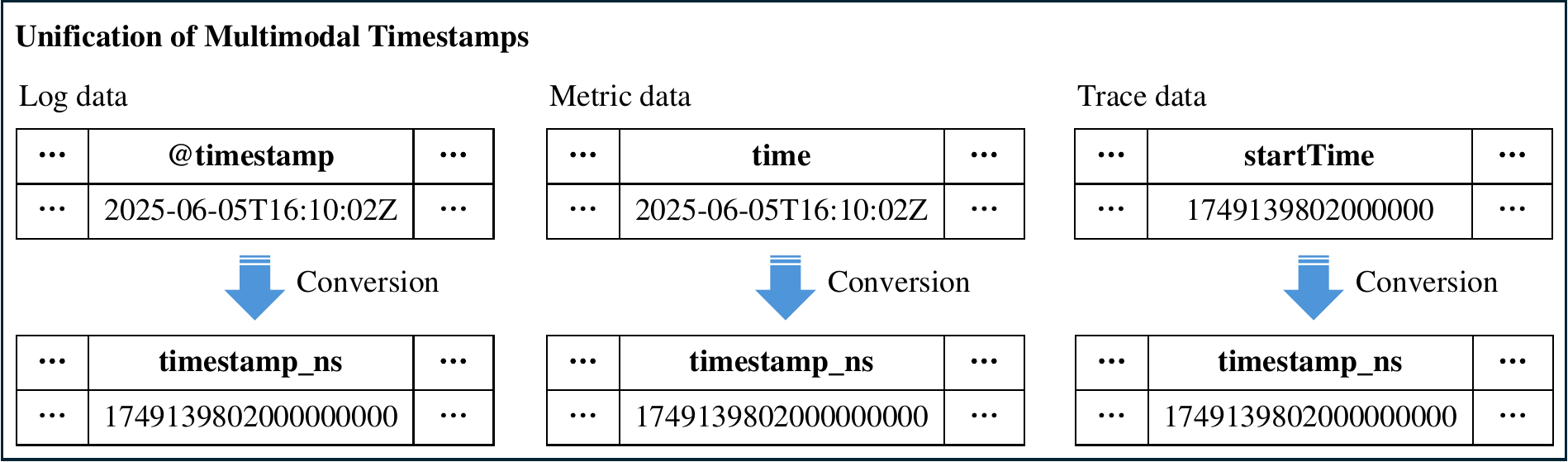}
    \caption{Unification of multimodal data timestamps.}
    \label{fig4}

    \vspace{10pt}
    \includegraphics[width=0.85\textwidth]{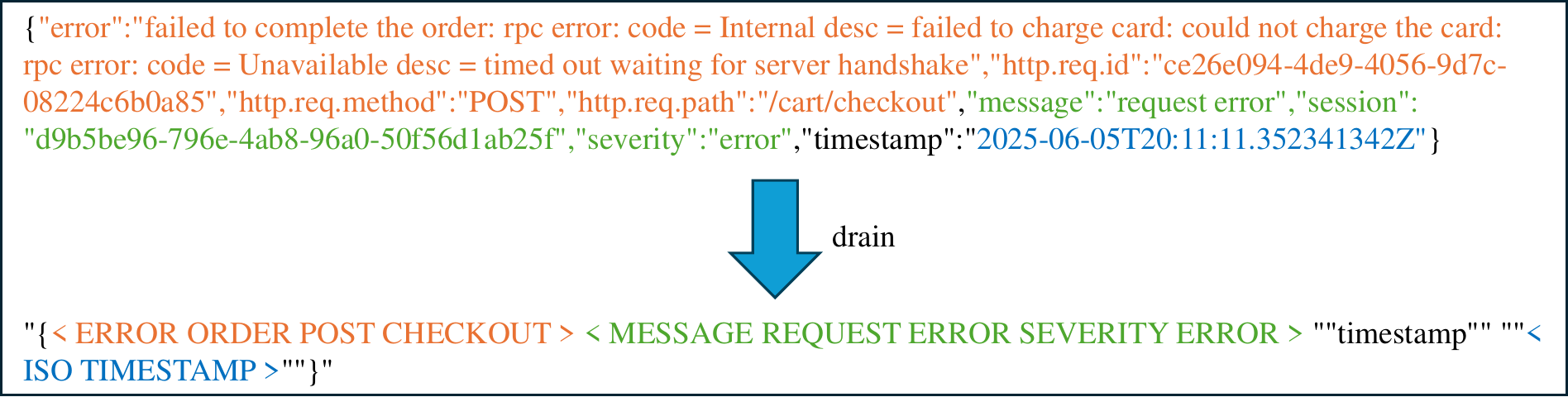}
    \caption{Log template extraction using Drain.}
    \label{fig5}
\end{figure*}

\begin{figure*}[!t]
    \centering
    \includegraphics[width=0.85\textwidth]{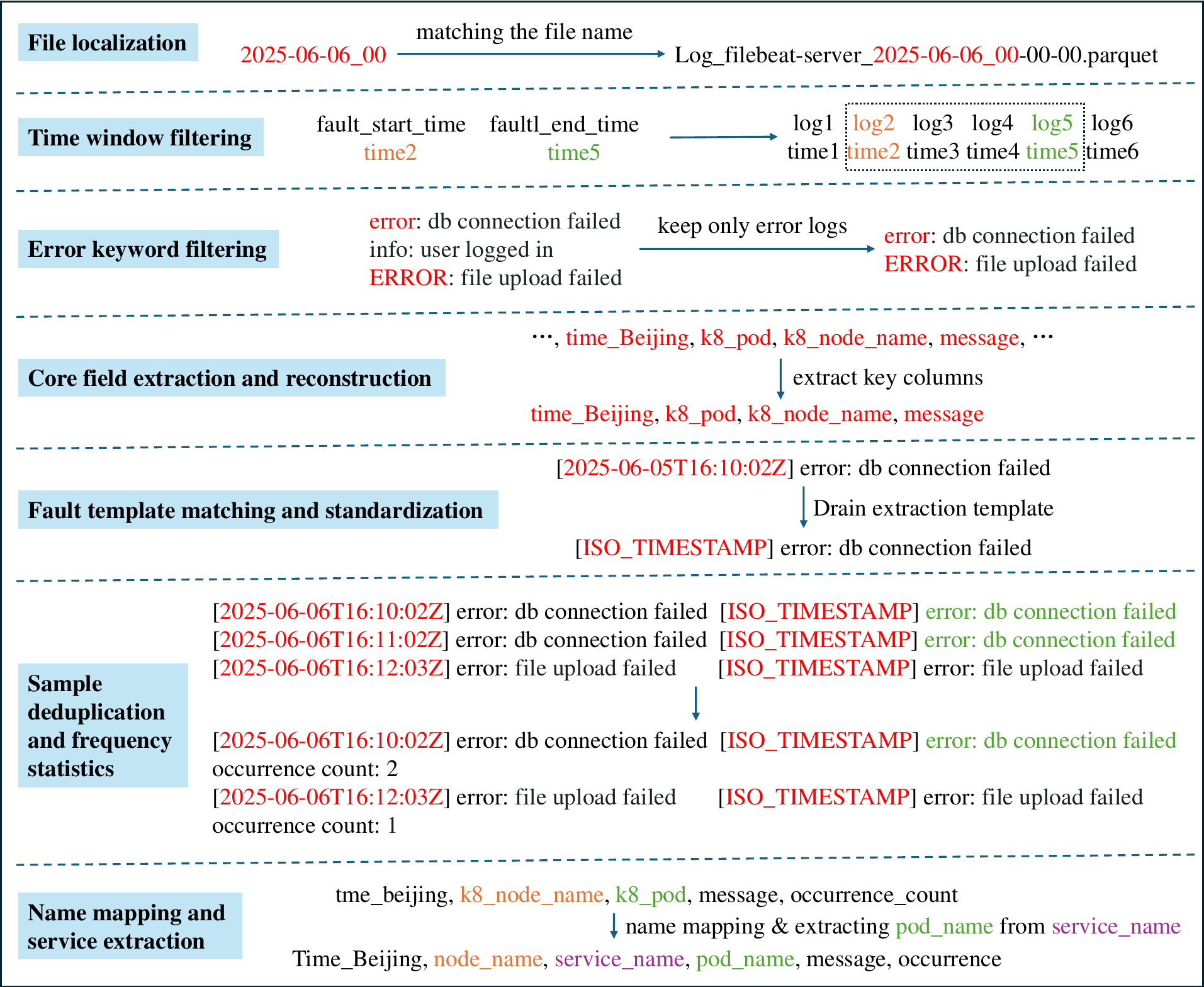}
    \caption{Multi-level data filtering and processing workflow.}
    \label{fig6}

    \vspace{10pt}
    \includegraphics[width=0.7\textwidth]{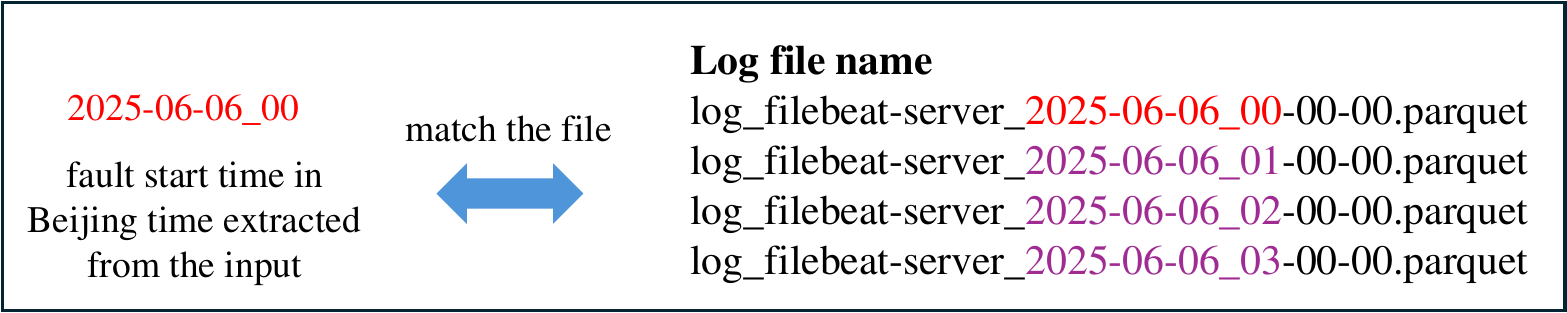}
    \caption{Matching between time in input and log file names.}
    \label{fig7}
\end{figure*}

\section{Detailed Design}
\label{sec:detail}

Centering on the demand for fault root cause localization in microservice systems, this solution constructs a complete system architecture consisting of five modules: data preprocessing, log fault extraction, trace fault detection, metric fault summarization, and multimodal root cause analysis (as shown in Figure \ref{fig1}). First, the data preprocessing module realizes the structured parsing of input fault information and the unification of multimodal timestamps. Subsequently, the log module performs multi-level data filtering on large-scale logs and uses the Drain algorithm to extract log templates for deduplication; the trace module adopts a dual strategy combining duration performance analysis and status verification to identify trace anomalies; the metric module leverages large language models to conduct phenomenon summarization for multi-level monitoring metrics. Finally, all modal data are aggregated into the multimodal root cause analysis module, which utilizes the cross-modal reasoning capability of large language models to achieve the localization of faulty components, the explanation of causes, and the output of evidence chains.

\subsection{Data Preprocessing}

The data preprocessing module focuses on two core functions: parsing input data and uniformly processing multimodal timestamps. It ensures the consistency and accuracy of subsequent processing for each modal data, thereby laying a data foundation for the entire fault root cause localization system.

\subsubsection{Design of Input Data Parsing}

For input data, a regex-based extraction strategy is adopted to perform structured processing on raw fault description files. The system accepts input data in JSON format, which contains two core components: an anomaly description and a unique identifier (uuid). Among these components, the most critical pieces of information are the fault start/end times (included in the anomaly description) and the uuid.

To facilitate accurate matching of subsequent multimodal data and filtering of data within the fault time window, the system first parses and extracts key information from the input files. The core information extracted includes the unique fault identifier and the fault start/end times, as illustrated in Figure \ref{fig2}.

\textbf{Timestamp Extraction Mechanism:} The system adopts the ISO 8601 time format standard and implements automated timestamp identification through the regular expression pattern \verb!(\d{4}-\d{2}-\d{2}T\d{2}:\d{2}:\d{2}Z)!. The extraction rule follows the strategy where the first match is taken as the fault start time and the second match as the fault end time.

\textbf{Time Index Generation:} To enhance the efficiency of subsequent data correlation, the system constructs a time indexing mechanism. It generates time identifiers in the format "year-month-day\_hour" (e.g., 2025-06-06\_00) for rapid localization of corresponding data files. Additionally, fault times are converted into 19-digit nanosecond-level timestamps, providing a high-precision reference for accurate time-range filtering. The preprocessed input data is specifically illustrated in Figure \ref{fig3}.

\subsubsection{Unification of Multimodal Data Timestamps}

In response to the distinct format characteristics of the three data types (log, trace, and metric), the system is designed with differentiated timestamp unification strategies in a systematic manner. These strategies enable the standardization of cross-modal time references, as specifically illustrated in Figure \ref{fig4}.

\textbf{Timestamp Standardization for Log Data:} For log data, the @timestamp field in ISO 8601 format is adopted as the time reference. The system parses the time format to convert raw timestamps into unified 19-digit nanosecond-level timestamps. After processing, the log data is sorted in ascending order of timestamps, ensuring the temporal consistency of the data and the accuracy of subsequent analyses.

\textbf{Timestamp Standardization for Metric Data:} Metric data stores time information in the "time" field, which also adheres to the ISO 8601 format standard. Given the multi-level distributed nature of metric data in the storage architecture, the system employs a recursive search strategy to traverse all subdirectory structures, ensuring complete coverage of metric files stored in a decentralized manner. The timestamp conversion logic remains consistent with that applied to log data.

\textbf{Timestamp Standardization for Trace Data:} Trace data features a unique time storage format, with microsecond-level timestamps stored in the "startTime" field. In this study, precision conversion is applied to extend microsecond-level timestamps to nanosecond-level ones (by multiplying by a factor of 1000), achieving time precision alignment with other modal data.

\subsection{Log Fault Extraction}

\begin{figure*}[htbp]
    \centering
    \includegraphics[width=0.75\textwidth]{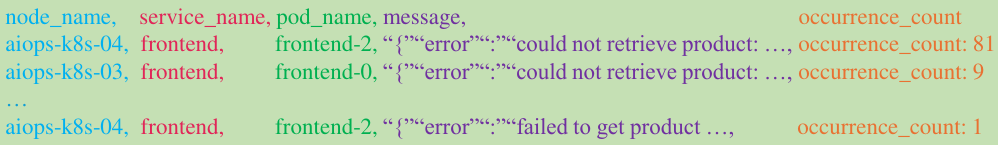}
    \caption{Output of log fault extraction.}
    \label{fig8}
\end{figure*}

The log fault extraction module serves as a core component of the fault root cause localization system, employing a multi-level filtering strategy to efficiently convert massive volumes of raw logs into structured fault feature information. The fault information compression functionality of this module is based on a pre-trained Drain algorithm model. It achieves effective compression of fault logs by extracting templates from logs containing error fields and performing deduplication on logs sharing the same template, thereby providing high-quality structured log data input for subsequent multimodal fault analysis.

\subsubsection{Log Parsing Principles and Model Construction}

\paragraph{Challenges in Unstructured Log Processing}

As unstructured information, logs record system operational behaviors in detail. However, they also contain a large number of irrelevant variables (e.g., timestamps, IP addresses, port numbers) that impair analysis efficiency. While these variables hold significance in logs, they tend to result in a multitude of duplicate records with identical semantics but different expressions during fault pattern recognition. Such duplicate records occupy valuable context space in large language models without providing additional valid information.

\paragraph{Selection and Application of Drain Algorithm}

After a comprehensive evaluation of existing log parsing techniques, this study adopts the widely recognized Drain algorithm as the core log parser. By constructing a parsing tree, the Drain algorithm can effectively identify fixed and variable parts within logs, and categorize logs with identical semantic patterns into the same template. This process enables the conversion of unstructured logs into structured features while compressing a large amount of redundant logs. Specifically, the Drain algorithm extracts log templates, removes irrelevant variables (e.g., timestamps, IDs), and merges logs sharing the same template into a single entry, thereby reducing the data volume effectively. Figure \ref{fig5} illustrates the log template extraction process: in its output, all variables are parsed into invariant content by Drain, which mitigates the interference caused by irrelevant variables.

\paragraph{Construction of the Pre-trained Model}

The system filters all log records containing the "error" field from the full-volume log data in phaseone\footnote{The dataset consists of two parts, phaseone and phasetwo; here we use the phaseone data for training. Source: \url{https://www.aiops.cn/gitlab/aiops-live-benchmark/phaseone/}.} to use as training samples, thereby constructing a pre-trained Drain model. Through the learning and analysis of massive error logs in the microservice environment, the model successfully extracts and builds 156 representative fault templates, which cover the main fault pattern types in the microservice system in a relatively comprehensive manner.

\subsubsection{Multi-level Data Filtering and Processing Workflow}

The multi-level data filtering and processing workflow is specifically illustrated in Figure \ref{fig6}, proceeding from file localization, through time window filtering, error keyword filtering, extraction and reconstruction of core fields, fault template matching and standardization, sample deduplication and frequency statistics, to name mapping and service extraction. This process ultimately converts massive volumes of logs into structured fault features while achieving effective content compression.

\paragraph{File Localization}

The system performs file matching using time identifiers in the "year-month-day\_hour" format, based on the time information of each input item output from the preprocessing stage, as illustrated in Figure \ref{fig7}. By searching for log files corresponding to the target time period within the project data directory, the system ensures accurate localization of data sources within the fault time window.

\paragraph{Time Window Filtering}

Based on the nanosecond-precision timestamps of the fault start and end times, strict time-boundary filtering is performed on log data. The system conducts a range query using the timestamp field, ensuring that only log records within the fault time period are retained and effectively eliminating interference from irrelevant information in the time.

\paragraph{Error Keyword Filtering}

\begin{figure*}[htbp]
    \centering
    \includegraphics[width=0.95\textwidth]{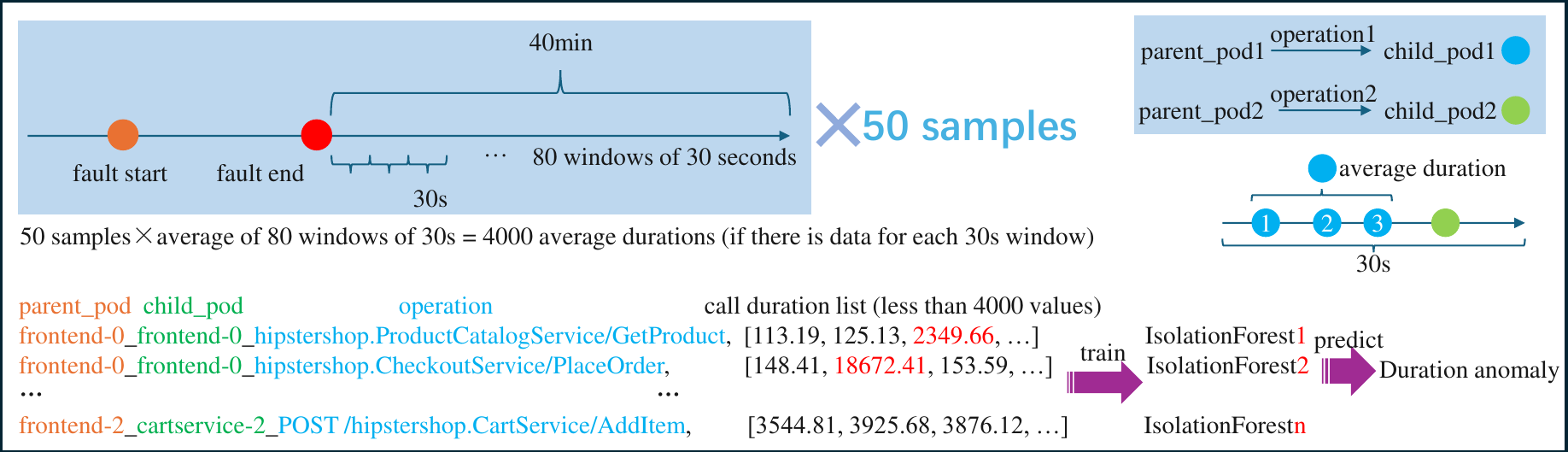}
    \caption{Construction of Isolation Forest training data and model training.}
    \label{fig9}
\end{figure*}

A filtering rule based on the "error" keyword is implemented for log message fields. After analyzing a large volume of log data, it is observed that faults are likely to occur in a large number of log entries accompanied by the "error" field. Therefore, log entries containing "error" information are automatically identified and extracted, while logs related to normal business operations are filtered out. This significantly enhances the relevance and efficiency of subsequent analyses.

\paragraph{Extraction and Reconstruction of Core Fields}

Key information essential for analysis is accurately extracted from multiple fields of raw logs, including critical dimensions such as time information (time\_beijing), container identifier (k8\_pod), node information (k8\_node\_name), and error message content (message). A field simplification strategy is employed to reduce data redundancy while ensuring information integrity.

\paragraph{Fault Template Matching and Standardization}

The pre-trained Drain model is utilized to perform template matching for each error log. The system converts raw log messages into standardized template representations, removes irrelevant variables, and retains only core fault semantic content, thereby enabling the extraction from individual logs to pattern categories.

\paragraph{Sample Deduplication and Frequency Statistics}

A deduplication strategy is applied to duplicate log records with the same container and template combination (e.g., [[2025-06-06 08:00, \textbf{frontend-0}, \textbf{template0}, message0], [2025-06-06 09:00, \textbf{frontend-0}, \textbf{template0}, message1]]). The system retains the first occurrence record of each combination and simultaneously counts the occurrence frequency of the combination (e.g., [2025-06-06 08:00, frontend-0, message0, Occurrence Count: 2]). This quantitative labeling provides an objective basis for evaluating the severity of faults.

\paragraph{Name Mapping and Service Extraction}

The k8\_node\_name and k8\_pod fields in logs are mapped and converted into node\_name and pod\_name respectively. Corresponding service information is then extracted by parsing the pod\_name (e.g., frontend-0 → frontend). Finally, the system reconstructs the data into a standardized multi-level format, which includes dimensions such as node, service, pod, error message, and frequency statistics. This provides a structured data foundation for subsequent multimodal fault analysis.

Through the processing workflow described above, the log fault extraction module can efficiently compress massive volumes of raw log data into a high-quality set of structured fault features. Specific examples of the finally extracted faults are illustrated in Figure \ref{fig8}. This module not only significantly reduces the complexity of data processing but also maintains the integrity and accuracy of fault information, thereby providing log-level data support for subsequent agent-based multimodal fault root cause localization.

\subsection{Trace Fault Detection}

The trace fault detection module is a critical component for fault root cause localization in microservice systems. It adopts a dual anomaly detection strategy to identify abnormal patterns in microservice traces from both performance and status dimensions. Based on the Isolation Forest algorithm and a direct status code inspection method, this module effectively detects duration anomalies and status anomalies. The two types of detection results are simultaneously fed into multimodal analysis, providing high-quality structured trace data input for subsequent multimodal fault analysis.

\subsubsection{Principle of Dual Anomaly Detection and Model Construction}

Distributed traces in microservice architectures contain abundant performance and status information, yet abnormal patterns exhibit diverse characteristics. Performance anomalies are primarily manifested as a significant deviation of trace duration from the normal range, which requires the establishment of a normal behavior baseline through machine learning methods. In contrast, status anomalies are characterized by the occurrence of explicit error codes and can be identified via direct inspection. The combination of these two anomaly detection mechanisms enables the provision of more comprehensive fault feature information.

\paragraph{Principle of Isolation Forest for Performance Anomaly Detection}

For detecting duration-based performance anomalies, the Isolation Forest unsupervised learning algorithm is employed. This algorithm operates on the core assumption that "anomalies are more easily isolated in the feature space" and quantifies the anomaly degree of data points by constructing multiple randomly partitioned trees. Isolation Forest is particularly suitable for anomaly detection in continuous numerical data, enabling automatic identification of significant deviation patterns in duration data without the need for labeled anomaly samples.

\paragraph{Direct Inspection Mechanism for Status Codes}

A direct inspection mechanism based on status codes is constructed, eliminating the need for a training process. The system directly analyzes the status.code and status.message fields within the tags of trace data, and filters out all abnormally statused invocations through a simple conditional judgment (\( \text{status.code} \neq 0 \)). This method can directly capture explicit abnormal states and specific information such as system errors, timeouts, and connection failures, providing deterministic anomaly evidence for fault root cause localization.

\paragraph{Construction of Isolation Forest Training Data and Model Training}

\begin{figure*}[htbp]
    \centering
    \includegraphics[width=1.0\textwidth]{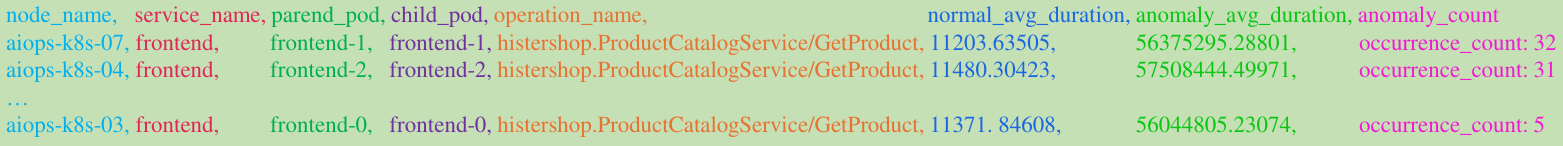}
    \caption{Example output of anomalous traces detected by Isolation Forest.}
    \label{fig10}
\end{figure*}

As illustrated in Figure \ref{fig9}, the construction of Isolation Forest training data and the subsequent model training primarily consist of the following components:

\textbf{Training Data Source:} In this scheme, data from the phaseone stage is used to construct the training set. Fifty fault samples are randomly selected, and a 40-minute time window immediately after the end of each fault is extracted as "normal period" data. This approach is based on the assumption that system behavior tends to return to normal after fault recovery, ensuring that the majority of the training data consists of normal data.

\textbf{Grouped Training Strategy:} Training data is grouped according to the service invocation combination of "parent\_pod - child\_pod - operation name", and an independent Isolation Forest detector is trained for each unique service invocation pattern. This fine-grained grouping strategy accounts for differences in performance baselines across various service invocations, thereby improving the accuracy of anomaly detection.

\textbf{Sliding Window Averaging Processing:} During the training phase, a 30-second sliding window is employed to process the duration data as an event sequence. Since multiple invocation combinations of "parent\_pod - child\_pod - operation name" may occur within a single 30-second sliding window, and each combination may appear one or more times, the average duration is calculated for each combination that occurs at least once within the window. If a combination does not appear, it is recorded as "None". This processing method effectively eliminates random fluctuations in individual invocation samples, stabilizes the duration feature, and thereby more accurately reflects the actual performance level of service invocations.

\textbf{Anomaly Detection Threshold Setting:} For the Isolation Forest model, the contamination parameter is set to 0.01 (indicating that the expected proportion of abnormal samples to the total is 1\%), and n\_estimators = 100 is used to construct 100 decision trees. The model outputs an anomaly score: a score of -1 indicates an abnormal sample, while a score of 1 indicates a normal sample.

\subsubsection{Multi-level Trace Data Processing Workflow}

\paragraph{File Localization and Time Filtering}

The system performs precise matching of trace files in the "year-month-day\_hour" format based on the time information of each input parsed in the preprocessing stage. Using nanosecond-level timestamps of the fault's start and end times, it executes time window filtering on the trace data, ensuring that the analysis scope covers the complete fault time period.

\paragraph{Invocation Relationship Mapping and Structured Extraction}

Key dimensional information is extracted from complex trace structures, including pod\_name, service\_name, and node\_name obtained by parsing the process field, as well as the establishment of complete parent-child relationship mappings for call chains using spanID and references. The system converts unstructured trace data into invocation representations containing "parent\_pod - child\_pod - operation name".

\paragraph{Duration Anomaly Detection Processing Flow}

\begin{figure*}[htbp]
    \centering
    \includegraphics[width=1.0\textwidth]{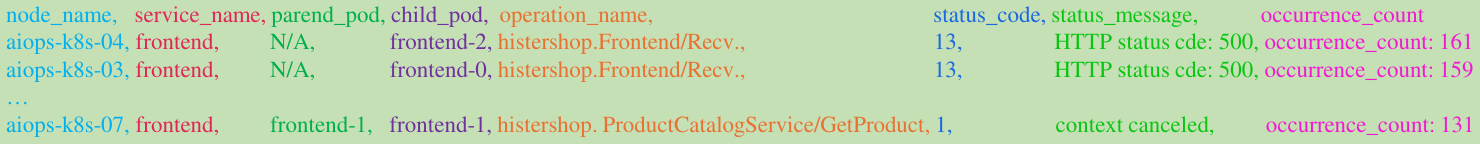}
    \caption{Example output of anomalous traces detected by status code.}
    \label{fig11}
\end{figure*}

\textbf{Prediction Data Processing:} Anomaly detection prediction is performed on all trace data within the fault time period, employing the same 30-second sliding window strategy as in the training phase to calculate the average duration within each time window. This consistent processing ensures the consistency of feature distribution between the training and prediction phases.

\textbf{Isolation Forest Anomaly Prediction:} The pre-trained detection model is used to identify anomalies in the duration features during the fault period. A dedicated detector corresponding to each service invocation combination is employed for prediction (if an invocation combination that was not trained before exists during the fault period, it is ignored and no prediction is performed), and anomaly labels are output (with -1 indicating an anomaly and 1 indicating normal).

\textbf{Duration Anomaly Result Output:} The system outputs the top 20 most frequent duration anomaly combinations, which include the following specific details:

\begin{itemize}
    \item node\_nameNode: identifier where the anomaly occurs;
    \item service\_name: Name of the anomalous service;
    \item parent\_pod: Identifier of the caller container;
    \item child\_pod: Identifier of the callee container;
    \item operation\_name: Name of the specific operation;
    \item normal\_avg\_duration: Average invocation latency during the normal period (derived from training data statistics);
    \item anomaly\_avg\_duration: Average invocation latency during the anomalous period (derived from actual statistics of the fault phase);
    \item anomaly\_count: Frequency of anomaly occurrences (format: Number of occurrences: N);
\end{itemize}

A specific example of the output of anomalous traces detected by the Isolation Forest is shown in Figure \ref{fig10}.

\paragraph{Status Anomaly Detection Processing Flow}

\textbf{Direct Status Check:} Status code checks are performed on all trace data within the fault time period, with no training process required. Information on status.code and status.message is parsed from the tags field of traces, and all anomalous status invocations are directly identified through conditional filtering (\( \text{status.code} \neq 0 \)).

\textbf{Anomalous Status Pattern Statistics:} The identified anomalous statuses are grouped and counted by service invocation combinations, and the occurrence frequency of each anomalous pattern is calculated.

\textbf{Status Anomaly Result Output:} The system outputs the top 20 most frequent status anomaly combinations, which include the following specific details:

\begin{itemize}
    \item Node\_name: Node identifier where the anomaly occurs;
    \item Service\_name: Name of the anomalous service (Redis is automatically mapped to redis-cart);
    \item Parent\_pod: Identifier of the caller container;
    \item Child\_pod: Identifier of the callee container;
    \item Operation\_name: Name of the specific operation;
    \item Status\_code: Specific error status code;
    \item Status\_message: Detailed description of the error status;
    \item Occurrence\_count: Frequency of status anomaly occurrences (format: Number of occurrences: N);
\end{itemize}

A specific example of the output of anomalous traces detected by the Status Check is shown in Figure \ref{fig11}.

Through the aforementioned dual anomaly detection workflow, the trace fault detection module achieves an organic combination of machine learning-based anomaly identification and rule-based direct anomaly checking. Duration anomaly detection provides quantitative performance anomaly analysis via the Isolation Forest algorithm, while status anomaly detection enables definitive error status identification through direct checks. The top 20 most critical anomaly patterns output by each of the two detection strategies collectively provide a trace data foundation for subsequent fault root cause localization.

\subsection{Metric Fault Summary}

\begin{figure}[htbp]
    \centering
    \includegraphics[width=0.45\textwidth]{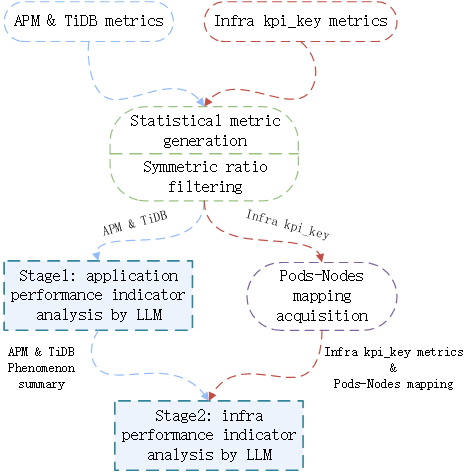}
    \caption{Schematic diagram of two-stage large language model phenomenon summary.}
    \label{fig12}
\end{figure}

The Metric Fault Summary Module serves as a full-stack monitoring and analysis component for faults in distributed microservice systems. It adopts a two-stage phenomenon summary strategy based on large language models (see Figure \ref{fig12}). By leveraging rule-based filtering and statistical comparison methods, it identifies significantly anomalous metrics. Furthermore, it utilizes the reasoning capabilities of large language models to perform phenomenon induction and pattern recognition on complex multi-dimensional monitoring data, thereby providing high-quality phenomenon description inputs for subsequent multimodal fault root cause analysis.

\subsubsection{Principle and Method Construction of Large Language Model-Driven Phenomenon Summary}

\paragraph{Limitations of Traditional Anomaly Detection Methods}

In distributed microservice architectures, monitoring metrics exhibit characteristics of high dimensionality, multi-level hierarchy, and strong correlations. Traditional anomaly detection methods, whether threshold-based or machine learning-based, often only identify numerical anomalies in individual metrics, struggling to capture the correlations between metrics and their business semantics. Furthermore, these methods lack the ability to generate semantic descriptions of anomalous phenomena, failing to provide interpretable phenomenon summaries for subsequent multimodal fault root cause localization.

\paragraph{Phenomenon Summary Method Based on Large Language Models}

In the metric data processing method of this solution, a large language model is employed as the core tool for phenomenon summary, fully leveraging its robust semantic understanding and reasoning-inductive capabilities. By inputting the filtered monitoring data, the model automatically identifies anomalous variation patterns, correlation relationships, and business impacts, and outputs phenomenon descriptions with rich semantic content. It is particularly emphasized that the large language model in this module is only used for phenomenon summary and pattern description, and does not perform fault judgment or root cause inference. This lays a foundation for subsequent comprehensive root cause analysis that integrates multimodal data such as logs and traces.

\paragraph{Definition of Normal Time Period}

The system defines the normal time period using a relative time window approach. The specific rules are as follows: the normal operation time periods are extracted as two intervals: one from 10 minutes after the end of the previous fault to the start of the current fault, and the other from 10 minutes after the end of the current fault to the start of the next fault. This definition method not only ensures that the data in the normal time period reflects the stable operation state of the microservice system (avoiding interference from the "aftermath" of faults) but also effectively reduces the impact of business fluctuations during different time periods of the day (e.g., high visit volume during the daytime and low visit volume at night) by selecting adjacent time windows before and after fault periods. In turn, it provides a more reliable performance baseline for comparative analysis.

\subsubsection{Multi-Level Monitoring Metric System and Filtering}

\paragraph{Pod-Service Unified Analysis Architecture}

The proposed system uniformly processes the pod and service levels, and automatically extracts service identifiers by parsing pod names. For instance, pod instances such as frontend-0, frontend-1, and frontend-2 are uniformly classified under the "frontend" service by removing their numeric suffixes. This design simplifies the analysis complexity in microservice architectures while maintaining dual perspectives at both the service level and instance level, enabling effective identification of service-level common issues and specific anomalies in individual instances.

\paragraph{Application Performance Monitoring Metric Filtering}

From the numerous redundant metrics in microservice APM monitoring data, 7 core key metrics are carefully selected: client\_error\_ratio (client error rate), error\_ratio (overall error rate), request (number of requests), response (number of responses), rrt (average response time), server\_error\_ratio (server error rate), and timeout (number of timeouts). These metrics cover the core dimensions of microservice performance, enabling comprehensive reflection of a service’s health status and anomalous characteristics while avoiding the impact of redundant data on the effectiveness of phenomenon analysis.

\paragraph{Hierarchical Coverage of Infra Performance Monitoring Metrics}

An infrastructure monitoring system covering both container and node layers is constructed. The container layer includes 9 pod infrastructure metrics: pod\_cpu\_usage (pod CPU usage rate), pod\_memory\_working\_set\_bytes (pod memory working set size), pod\_fs\_reads\_bytes (pod filesystem read bytes), pod\_fs\_writes\_bytes (pod filesystem written bytes), pod\_network\_receive\_bytes (pod network received bytes), pod\_network\_receive\_packets (pod network received packets), pod\_network\_transmit\_bytes (pod network transmitted bytes), pod\_network\_transmit\_packets (pod network transmitted packets), and pod\_processes (number of running processes in pod). The node layer comprises 16 node infrastructure metrics: node\_cpu\_usage\_rate (node CPU usage rate), node\_memory\_usage\_rate (node memory usage rate), node\_disk\_read\_bytes\_total (total disk read bytes), node\_disk\_written\_bytes\_total (total disk written bytes), node\_disk\_read\_time\_seconds\_total (total disk read time in seconds), node\_disk\_write\_time\_seconds\_total (total disk write time in seconds), node\_filesystem\_usage\_rate (filesystem usage rate), node\_network\_receive\_bytes\_total (total network received bytes), node\_network\_transmit\_bytes\_total (total network transmitted bytes), node\_network\_receive\_packets\_total (total network received packets), node\_network\_transmit\_packets\_total (total network transmitted packets), and node\_sockstat\_TCP\_inuse (number of TCP connections), among others. Hierarchical monitoring ensures comprehensive coverage from microservice applications to the underlying infrastructure of specific deployments.

\paragraph{Monitoring Metrics for TiDB Database Components}

A specialized monitoring metric system is constructed for the TiDB distributed database, covering three core components: tidb-tidb, tidb-tikv, and tidb-pd. It includes database-specific metrics such as failed\_query\_ops (number of failed queries), duration\_99th (99th percentile request latency), connection\_count (number of connections), server\_is\_up (number of alive service nodes), cpu\_usage (CPU usage rate), memory\_usage (memory usage), store\_up\_count (number of healthy Stores), store\_down\_count (number of down Stores), store\_unhealth\_count (number of unhealthy Stores), storage\_used\_ratio (used capacity ratio), available\_size (available storage capacity), raft\_propose\_wait (RaftPropose waiting latency), raft\_apply\_wait (RaftApply waiting latency), and rocksdb\_write\_stall (number of RocksDB write stalls). This enables professional monitoring and analysis of the data storage layer.

\subsubsection{Data Processing and Anomaly Filtering Workflow}

\paragraph{Hierarchical Data Loading and Organization}

\begin{figure*}[htbp]
    \centering
    \includegraphics[width=1.0\textwidth]{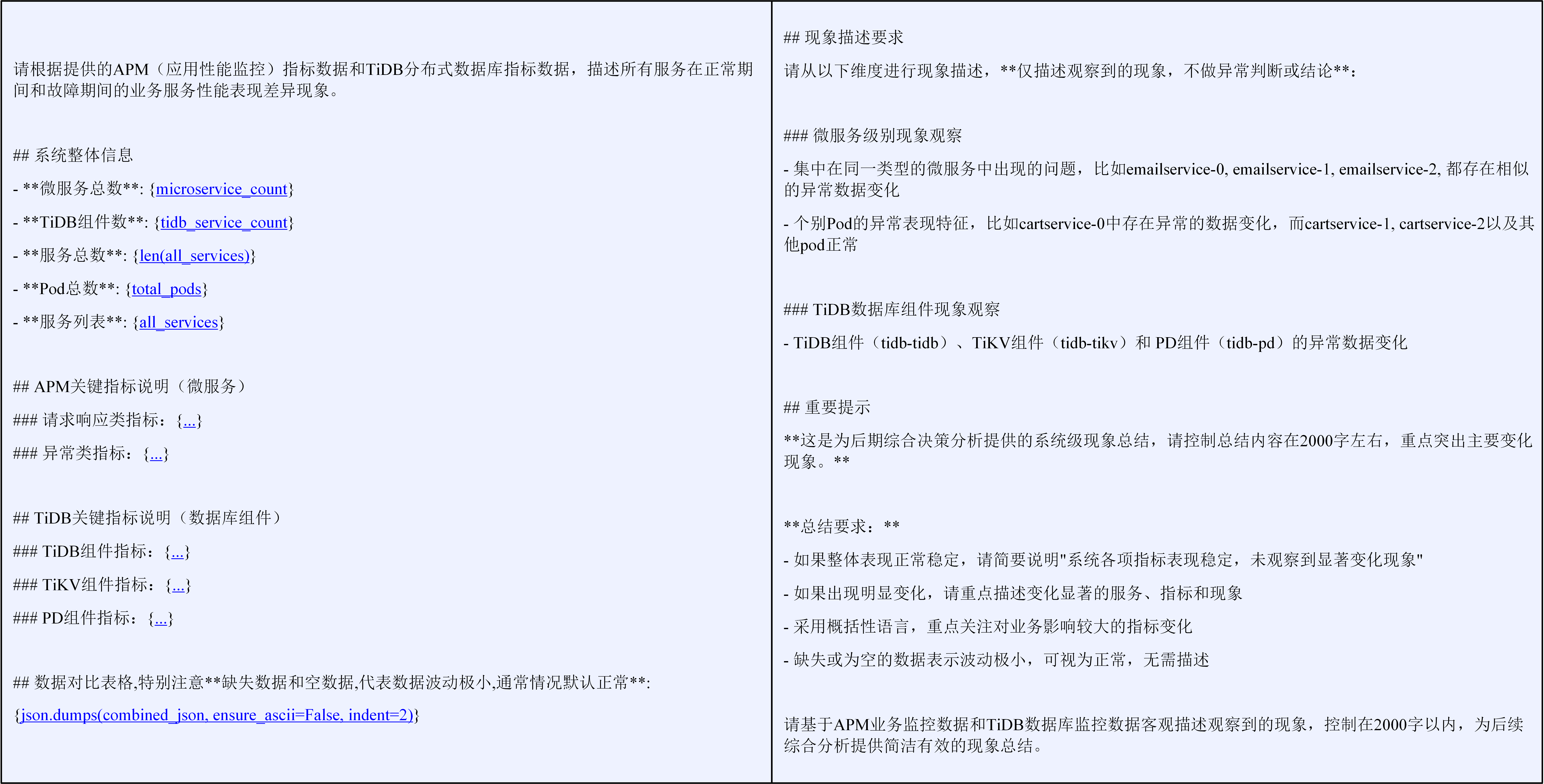}
    \caption{Prompt for stage 1's comprehensive phenomenon summarization (The prompt is designed in Chinese; see source code and modify as needed).}
    \label{fig13}
\end{figure*}

\begin{figure*}[!t]
    \centering
    \includegraphics[width=0.75\textwidth]{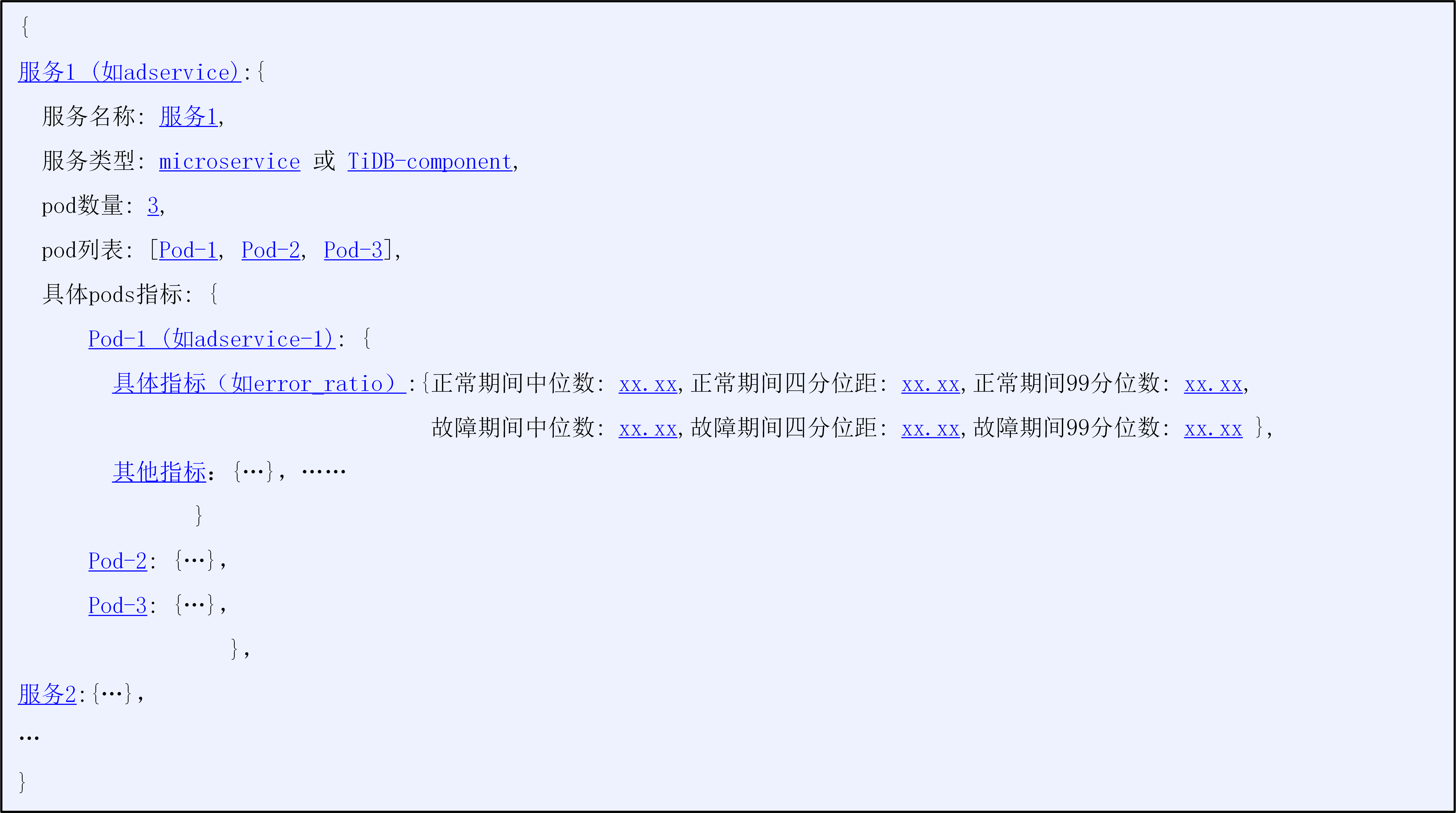}
    \caption{Example of comparative data for business metrics and TiDB database components.}
    \label{fig14}
\end{figure*}

The system constructs distinct data paths for file localization based on monitoring levels, using each input date information output from the preprocessing stage. Specifically, APM business monitoring data is stored in the directory "Year-Month-Day/metric-parquet/apm/pod", infrastructure node monitoring data resides in "Year-Month-Day/metric-parquet/infra/infra\_node", container monitoring data is located in "Year-Month-Day/metric-parquet/infra/infra\_pod", and TiDB database monitoring data is stored in directories such as "Year-Month-Day/metric-parquet/infra/infra\_tidb". Through this hierarchical file organization method, precise data localization and loading are realized in accordance with monitoring dimensions.

\paragraph{Precise Time Window Filtering and Data Merging}

The system performs nanosecond-level time window filtering on monitoring data at all levels to ensure that the analysis scope fully covers the fault time period. Data from two normal time periods before and after the fault period are merged. The 10 minutes immediately preceding the fault period are excluded to eliminate the impact of fault "aftermath", and random fluctuation effects are removed by excluding the two maximum and two minimum extreme values, thereby constructing a stable statistical baseline. This processing method ensures the representativeness and stability of data during normal periods.

\paragraph{Significance Filtering Based on Statistical Symmetric Ratio}

A significance filtering mechanism based on statistical symmetric ratios is implemented, which calculates the symmetric ratios of medians and 99th percentiles between the fault period and normal periods. Stable metrics with a variation amplitude of less than 5\% are automatically filtered out, while only potentially critical abnormal metrics with significant changes are retained for the large language model analysis process. This filtering strategy can compress massive raw data into a set of high-value abnormal features, substantially reducing the context of the large language model by approximately 50\% in terms of token count. Meanwhile, it significantly improves the efficiency and quality of phenomenon summarization. Taking the P50 symmetric ratio as an example, its specific formula is defined as follows:

\begin{equation}
P50_{\mathrm{symmetric\text{-}ratio}} = \frac{ \left| P50_{\mathrm{fault}} - P50_{\mathrm{normal}} \right| }{ \frac{ P50_{\mathrm{fault}} + P50_{\mathrm{normal}} }{2} + \varepsilon }
\end{equation}
where $P50$ denotes the median, $P50_{\mathrm{fault}}$ and $P50_{\mathrm{normal}}$ represent the medians during the fault period and normal periods respectively, and $\varepsilon$ is a minimal value.

\begin{figure*}[!t]
    \centering
    \includegraphics[width=0.6\textwidth]{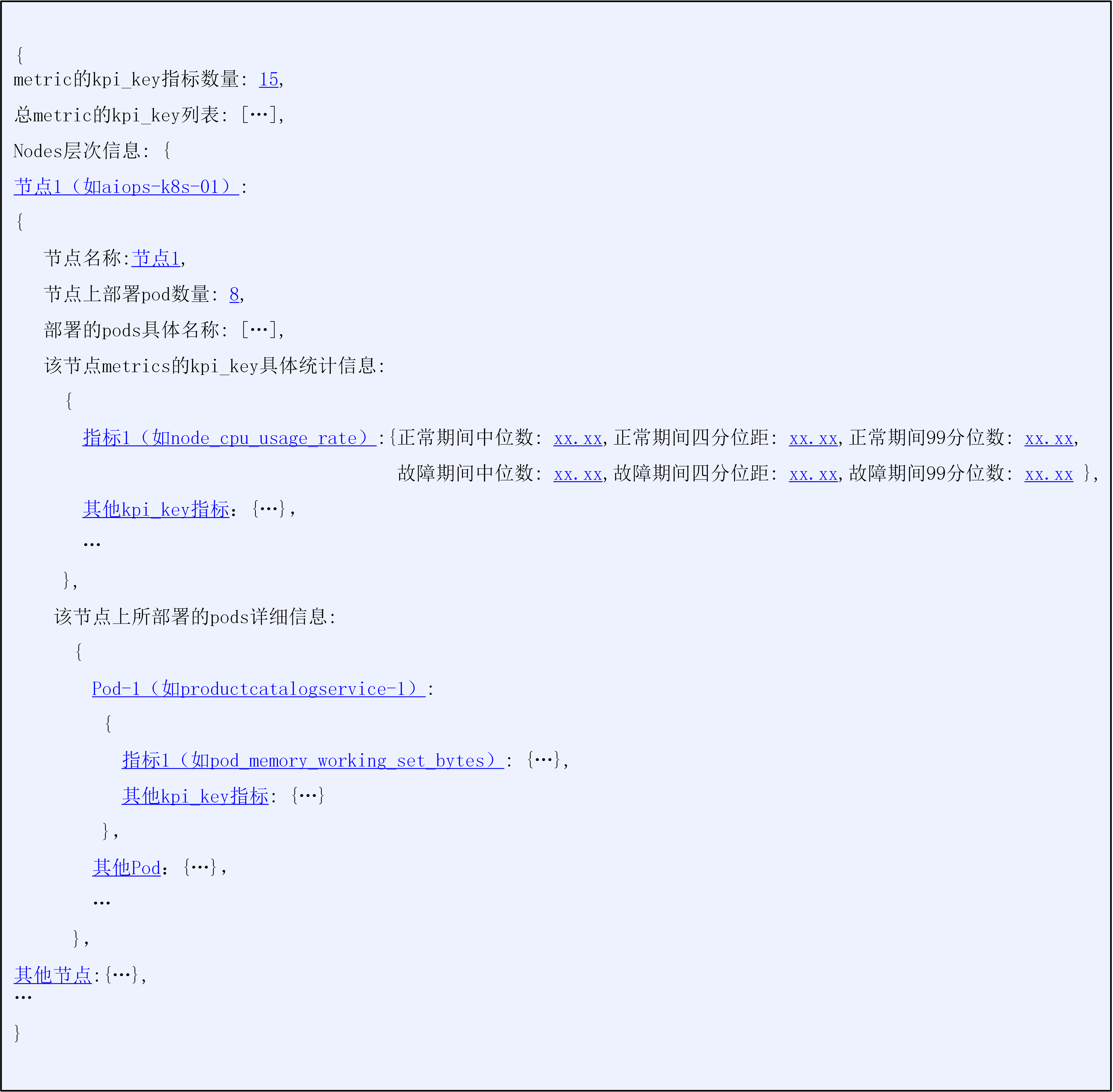}
    \caption{Example of statistical data comparison for kpi\_key metrics between different nodes and the pods deployed on them.}
    \label{fig15}
\end{figure*}

\begin{figure*}[!t]
    \centering
    \includegraphics[width=0.9\textwidth]{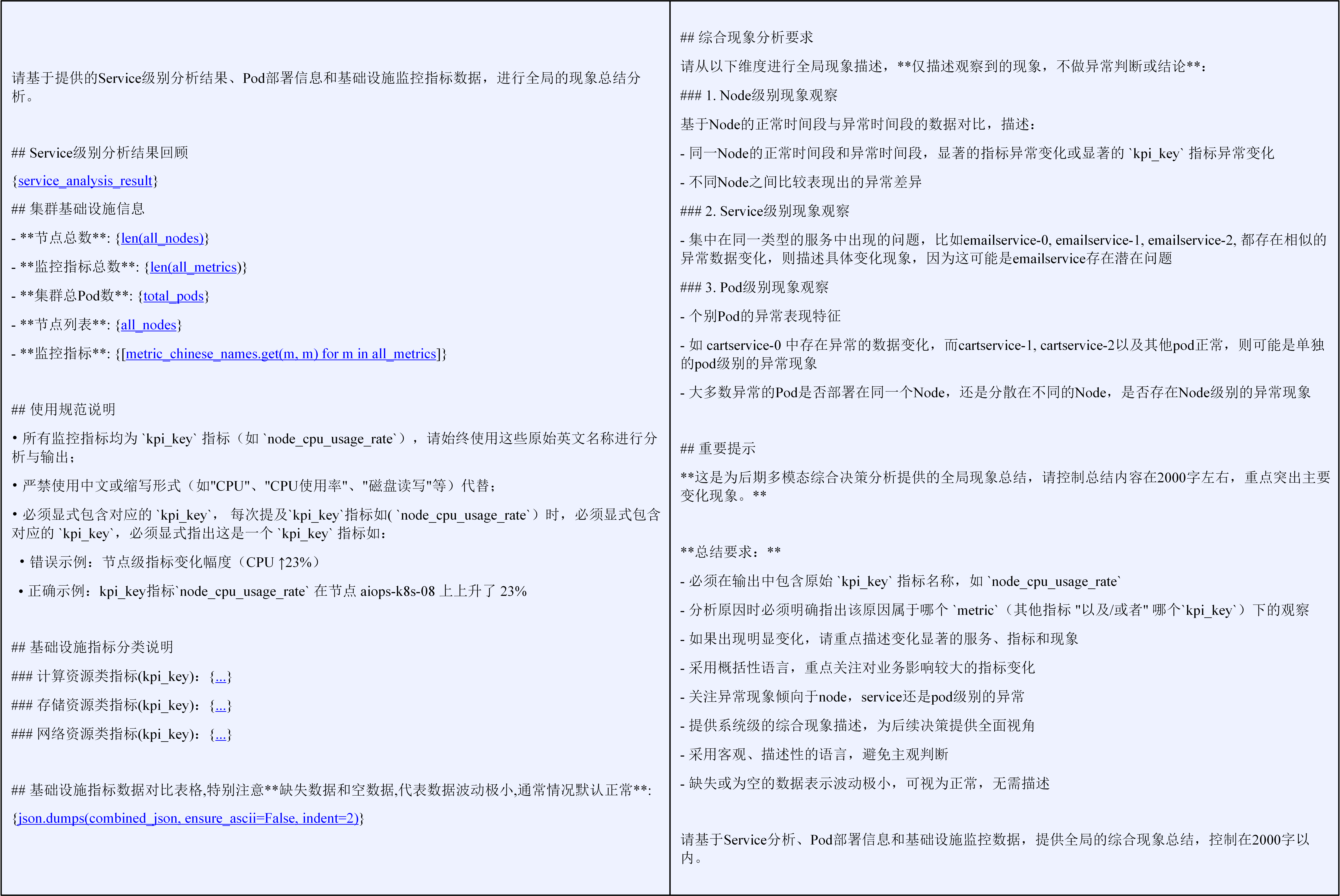}
    \caption{Prompt for the stage 2's comprehensive phenomenon summarization.}
    \label{fig16}
\end{figure*}

For the filtered key metrics, the pandas `describe` method is utilized to extract comprehensive descriptive statistical information, including key statistics such as median, interquartile range, and 99th percentile. Through a standardized data representation format, monitoring metrics of different types and dimensions are converted into structured inputs understandable to large language models, ensuring the accuracy and consistency of subsequent phenomenon summarization.

\subsubsection{Two-Stage Large Language Model Phenomenon Summarization and Output Generation}

\paragraph{Stage 1: Application Performance Monitoring Phenomenon Identification and Summarization}

The system first conducts a comprehensive analysis of microservice application performance monitoring metrics and TiDB database component metrics, organizing filtered key abnormal metrics by service type where each service includes data from its multiple subordinate Pod instances; for example, the emailservice contains comparative APM metrics such as client\_error\_ratio (client error rate), error\_ratio (overall error rate), request (number of requests), response (number of responses), rrt (average response time), server\_error\_ratio (server error rate), and timeout (number of timeouts) across Pod instances like emailservice-0, emailservice-1, and emailservice-2, while integrating database-specific metrics of TiDB components including failed\_query\_ops (number of failed queries), duration\_99th (99th percentile request latency), and connection\_count (number of connections). A structured table with comparative data between normal periods and the fault period is constructed, and using a service-Pod hierarchical organization method, the large language model is invoked for the first-round phenomenon summarization (with prompt design shown in Figure \ref{fig13}); the data comparison table adopts JSON format, which, compared with Markdown, maintains clarity while significantly reducing token usage (specific format presented in Figure \ref{fig14}), and while analyzing Pod-level phenomena, the system naturally summarizes service-level common phenomena and differentiated performances, with the large language model only describing phenomena at this stage without making fault judgments.

\paragraph{Stage 2: Comprehensive Phenomenon Summarization and Correlation Analysis of Infrastructure Machine Performance Metrics}

\begin{figure*}[!t]
    \centering
    \includegraphics[width=0.95\textwidth]{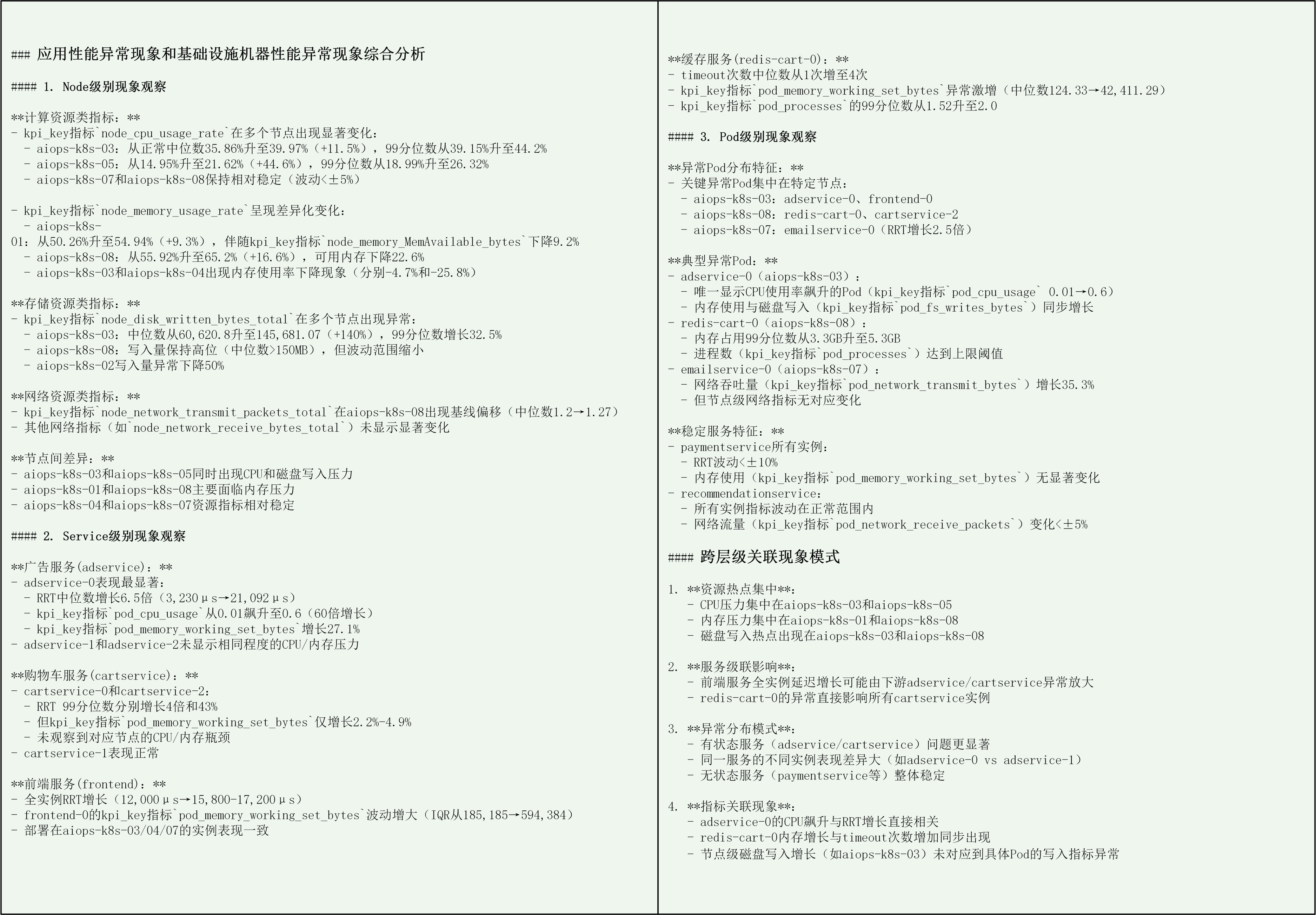}
    \caption{Output example of the metric fault summarization module.}
    \label{fig17}
\end{figure*}

Building on the analysis of application performance monitoring data, the system further integrates machine performance monitoring data from the infrastructure layer, encompassing both the pod container layer and node layer, and by combining the deployment topology information of pods on each node with the service analysis results from Stage 1, it constructs a full-stack phenomenon view covering three levels—application, container, and node—integrating container metrics such as pod\_cpu\_usage (Pod CPU usage rate), pod\_memory\_working\_set\_bytes (Pod memory working set size), and pod\_network\_receive\_bytes (Pod network received bytes), as well as node metrics including node\_cpu\_usage\_rate (node CPU usage rate), node\_memory\_usage\_rate (node memory usage rate), node\_disk\_read\_bytes\_total (total disk read bytes), and node\_network\_transmit\_bytes\_total (total network transmitted bytes); the large language model is then invoked to conduct the second-round comprehensive phenomenon summarization, with data comparison tables also adopting JSON format that, after describing the statistical indicators of node-level kpi\_key, proceeds to describe the kpi\_key indicators related to pods deployed on that node to ensure a coherent data flow (specific format shown in Figure \ref{fig15} and prompt design presented in Figure \ref{fig16}), emphasizing the analysis of cross-level abnormal correlation relationships to output a complete fault phenomenon description spanning from microservice applications to infrastructure, while this stage also only performs phenomenon summarization to lay the foundation for subsequent multimodal root cause analysis.

\paragraph{Structured Phenomenon Output}

\begin{figure*}[!t]
    \centering
    \includegraphics[width=0.95\textwidth]{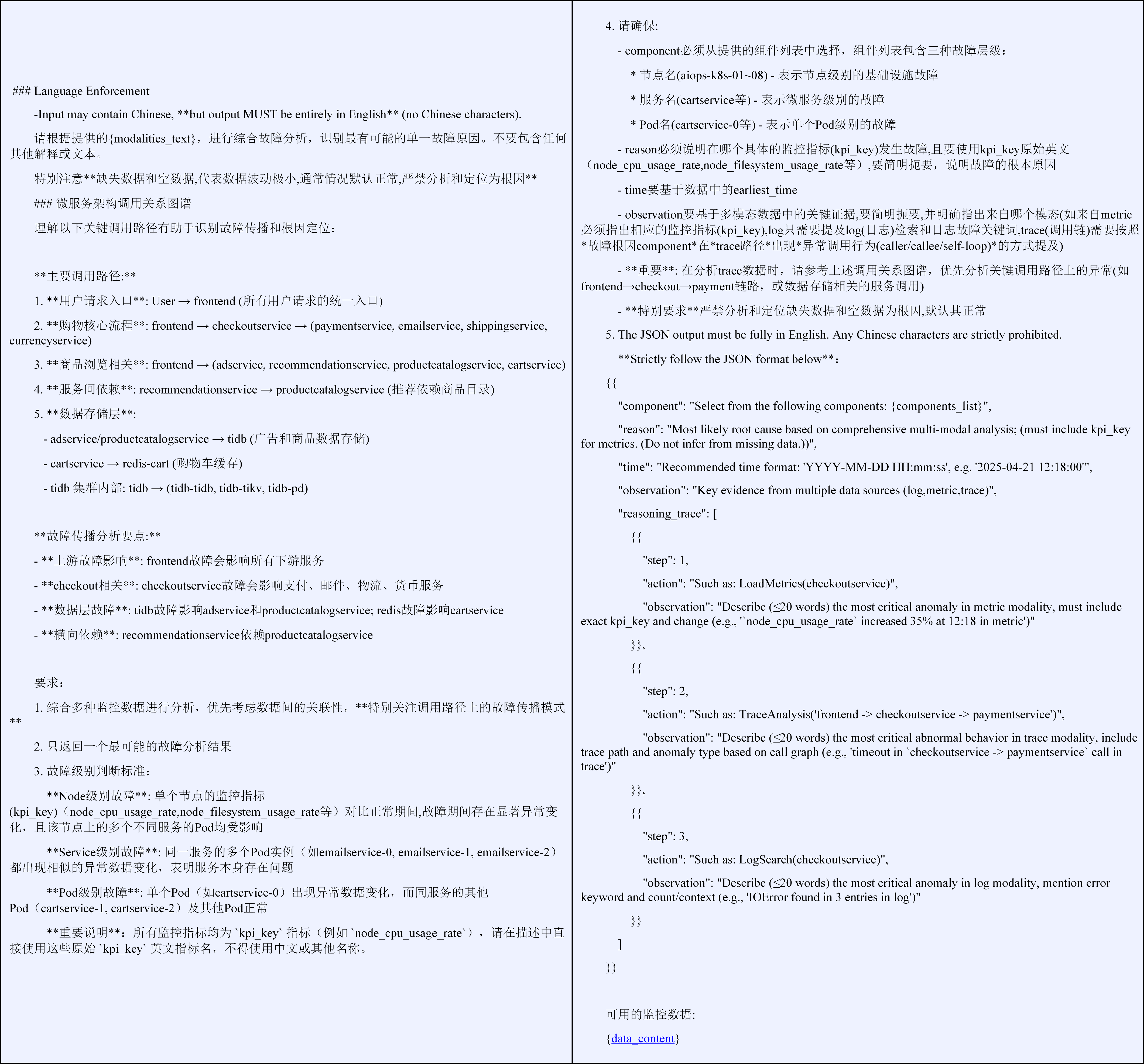}
    \caption{Prompt of multimodal analysis.}
    \label{fig18}
\end{figure*}

After processing via the two-stage large language model analysis, the metric fault summarization module outputs a phenomenon description that includes the following content:

\begin{figure*}[!t]
    \centering
    \includegraphics[width=0.65\textwidth]{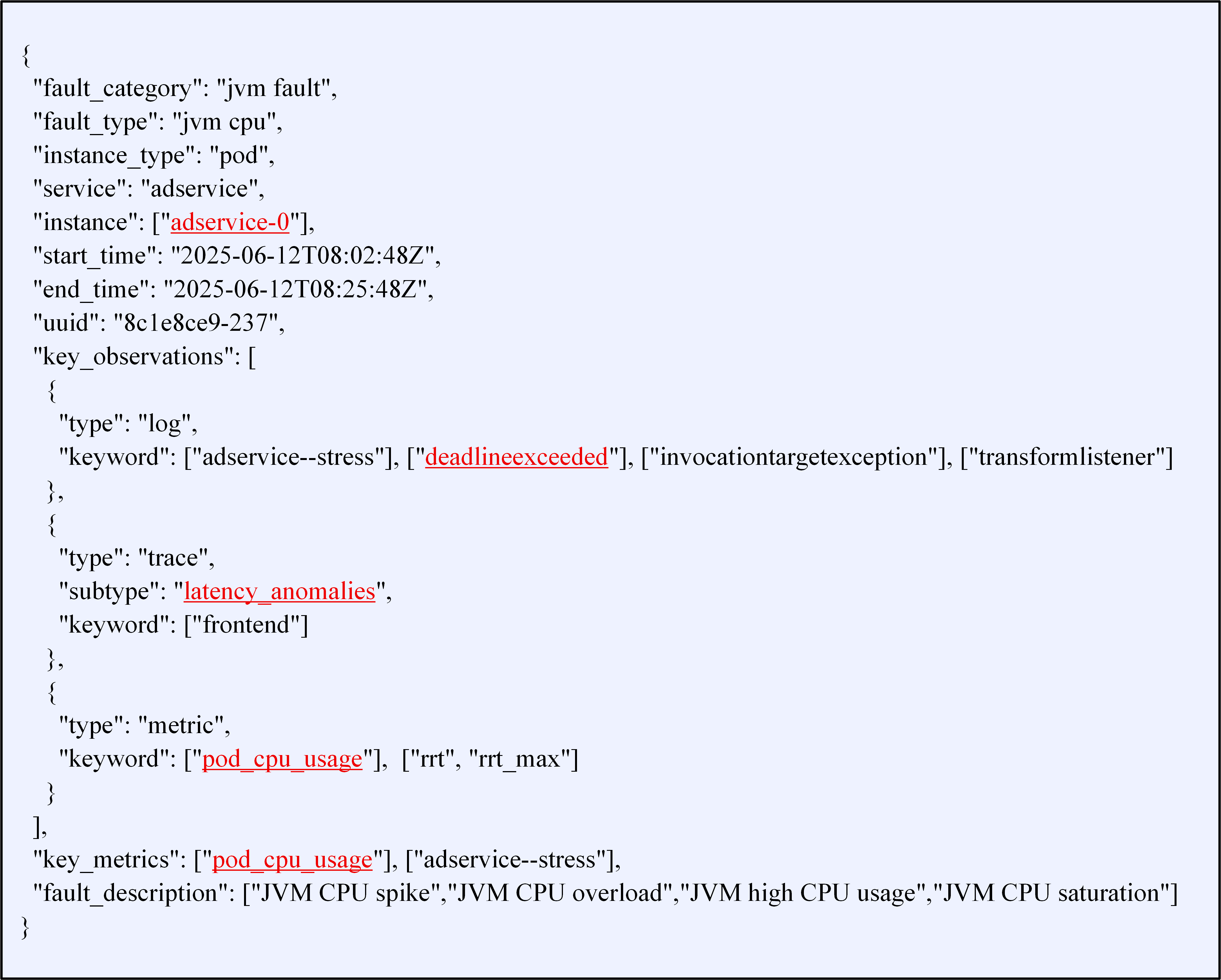}
    \caption{Good case - ground truth range.}
    \label{fig19}
\end{figure*}

\begin{figure*}[!t]
    \centering
    \includegraphics[width=0.7\textwidth]{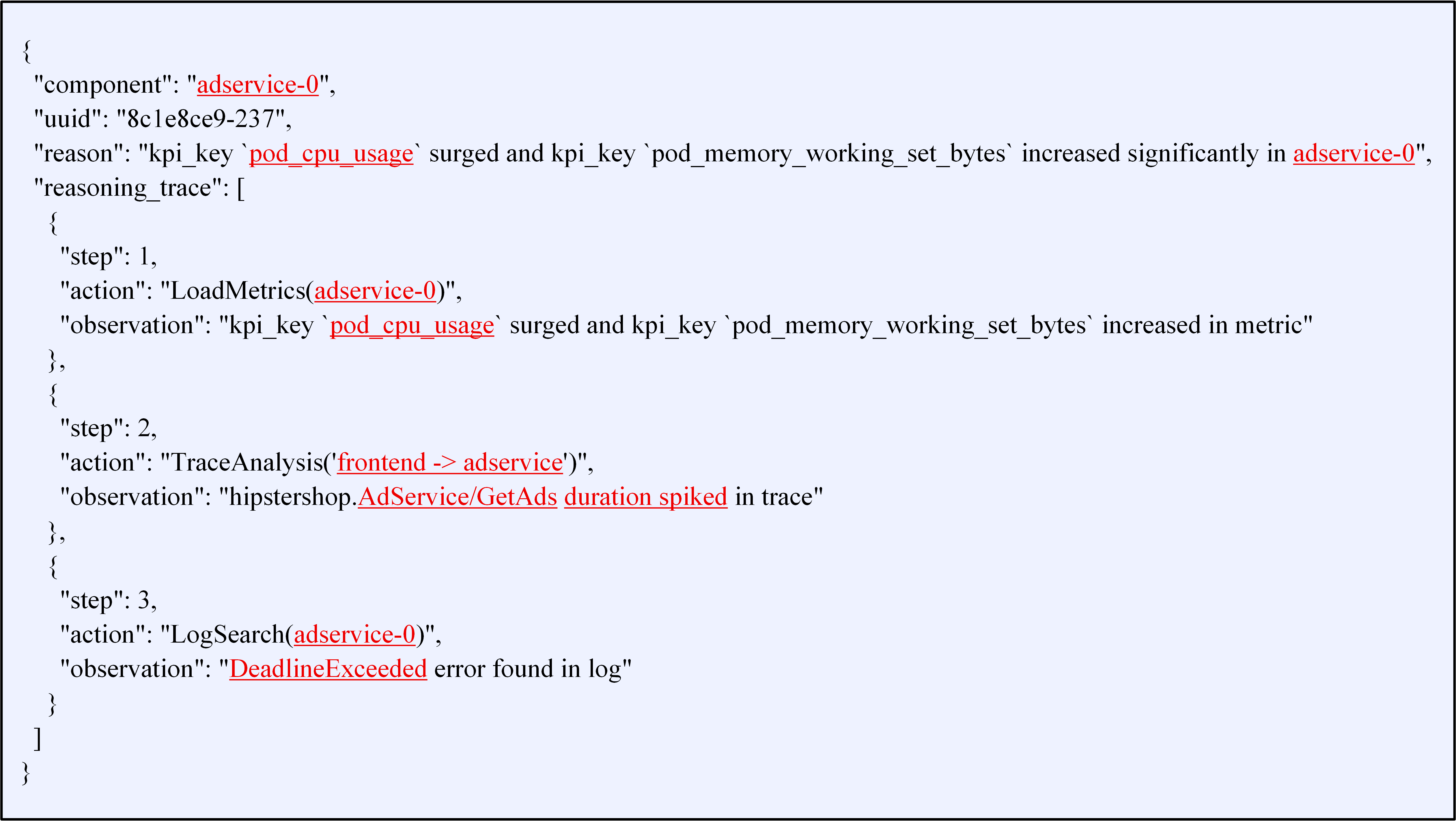}
    \caption{Good case - actual output values.}
    \label{fig20}
\end{figure*}

\textbf{Application Performance Anomaly Phenomena:} Leveraging the service-Pod hierarchical information, microservice components and database components with significant performance changes are identified. This includes changes in request processing metrics such as request (number of requests), response (number of responses), and rrt (average response time); fluctuations in abnormal metrics like error\_ratio (overall error rate), client\_error\_ratio (client error rate), server\_error\_ratio (server error rate), and timeout (number of timeouts); and variations in database performance metrics of TiDB components, including failed\_query\_ops (number of failed queries), duration\_99th (99th percentile request latency), and connection\_count (number of connections). This enables the simultaneous representation of overall trends at the service level and individual differences at the Pod level.

\textbf{Infrastructure Machine Performance Anomaly Phenomena:} This section describes changes in resource status at the container and node levels, including variation patterns of computing resource metrics such as node\_cpu\_usage\_rate (node CPU usage rate) and pod\_cpu\_usage (pod CPU usage rate); memory resource metrics like node\_memory\_usage\_rate (node memory usage rate) and pod\_memory\_working\_set\_bytes (pod memory working set size); storage resource metrics including node\_disk\_read\_bytes\_total (total disk read bytes), node\_disk\_written\_bytes\_total (total disk written bytes), and pod\_fs\_reads\_bytes (pod file system read bytes); and network resource metrics such as node\_network\_receive\_bytes\_total (total network received bytes), node\_network\_transmit\_bytes\_total (total network transmitted bytes), and pod\_network\_receive\_bytes (pod network received bytes), as well as spatial characteristics of anomalies distributed across different nodes and containers.

\textbf{Cross-Level Correlated Phenomenon Patterns:} Through Pod deployment topology and phenomenon correlation analysis, the correlation patterns between service anomalies and infrastructure anomalies are identified, and the distribution characteristics and propagation paths of abnormal phenomena are determined. This provides critical phenomenon clues for subsequent multimodal fault root cause analysis that integrates log and trace data.

Through the systematic processing workflow outlined above, the metric fault summarization module achieves the conversion of massive multi-dimensional monitoring data into concise phenomenon descriptions. By fully leveraging the semantic understanding and reasoning capabilities of large language models, this module focuses on phenomenon summarization rather than fault judgment. It not only significantly improves the accuracy and completeness of abnormal phenomenon identification but also provides a high-quality semanticized metric data foundation for subsequent fault root cause analysis that integrates multimodal data.

The output example of the metric fault summarization module is specifically shown in Figure \ref{fig17}.

\subsection{Multimodal Root Cause Analysis}

The multimodal root cause analysis module serves as the core decision-making component for fault root cause localization in microservice systems. It adopts a comprehensive reasoning strategy based on large language models, conducting integrated analysis of preprocessed fault log data, abnormal trace patterns, and summarized metric phenomena to achieve automated reasoning from multi-dimensional abnormal information to accurate root cause localization. By fully leveraging the cross-modal understanding and logical reasoning capabilities of large language models, this module provides final component localization and cause explanation for complex microservice faults.

\subsubsection{Input Data Integration}

The system first integrates the processing results of the three modalities. The log fault extraction module, based on the Drain log parsing algorithm combined with a multi-level data filtering mechanism, compresses massive raw logs into high-quality fault features. The trace fault detection module outputs two types of abnormal information: the duration anomaly detection results include the top 20 most frequent anomalies identified by Isolation Forest, covering call relationships, performance comparisons, and anomaly statistics; the status anomaly detection results include the top 20 most frequent status code anomalies, covering error types and specific descriptions. The metric fault summarization module outputs a phenomenon description summary analyzed by the two-stage large language model, which contains approximately 2,000 words of detailed abnormal phenomenon descriptions, covering monitoring anomaly patterns across node, service, and pod levels.

\subsubsection{Multimodal Prompt Generation}

Based on the integrated multimodal data, the system constructs a dedicated cross-modal analysis prompt template. The prompt design adheres to a structured principle, clearly identifying the source, type, and analysis focus of data from each modality to provide clear data understanding guidance for the large language model. Meanwhile, the output requirements and format specifications are explicitly defined in the prompt, ensuring the large language model can conduct systematic root cause reasoning based on the multimodal evidence chain. The prompt design is shown in Figure \ref{fig18}.

\subsubsection{Structured Output Format Design}

The system primarily guides the standardized output format by providing output example prompts, ensuring the consistency and interpretability of root cause analysis results. The output result includes three core fields: Fault Component (component), Fault Reason (reason), and Reasoning Process (reasoning\_trace). The Fault Component field clearly identifies the name of the problematic microservice component; the Fault Reason field provides a concise and clear root cause description; and the Reasoning Process field details the model’s analytical logic and evidence chain, laying the foundation for result verification and subsequent optimization.

\subsubsection{Result Extraction and Verification Mechanism}

Due to the potential instability of large language model outputs (such as missing closing brackets, inclusion of redundant explanations, etc.), the system employs regular expressions and structured parsing methods to extract standardized results from the natural language outputs of large language models. It ensures the format correctness of output results through JSON format verification and guarantees the validity of key fields via content integrity checks. For results with parsing failures or format inconsistencies, the system provides retry and exception handling procedures to maximize the quality of outputs.

Through the multimodal root cause analysis workflow outlined above, the system achieves end-to-end automated processing from heterogeneous monitoring data to fault localization. By fully leveraging the advantages of large language models in cross-modal understanding and logical reasoning, this module provides an efficient, accurate, and interpretable solution for fault root cause localization in complex microservice systems.

\section{Result Analysis}

\subsection{Comparison Between Good Cases and Bad Cases}

As shown in Figure \ref{fig19} and Figure \ref{fig20}, the proposed intelligent fault root cause localization system with multimodal data fusion can accurately identify the most likely faulty components and key root cause derivation evidence from the three modalities of monitoring data (logs, trace data, and system metrics) through in-depth integration. This demonstrates the effective implementation of the designed multimodal extraction scheme. Furthermore, the scheme is capable of making reasonable sequential reasoning, effectively identifying fault hierarchy types, and determining the actual hierarchical sources (pods, services, nodes) of potentially abnormal data indicators. Ultimately, the multimodal root cause analysis module can make reasonable fault root cause judgments based on log fault data, trace anomaly patterns, and metric phenomena. By effectively combining the multimodal analysis and logical reasoning capabilities of large language models, it achieves the goal of reasonable fault root cause localization.

\begin{figure*}[!t]
    \centering
    \includegraphics[width=0.75\textwidth]{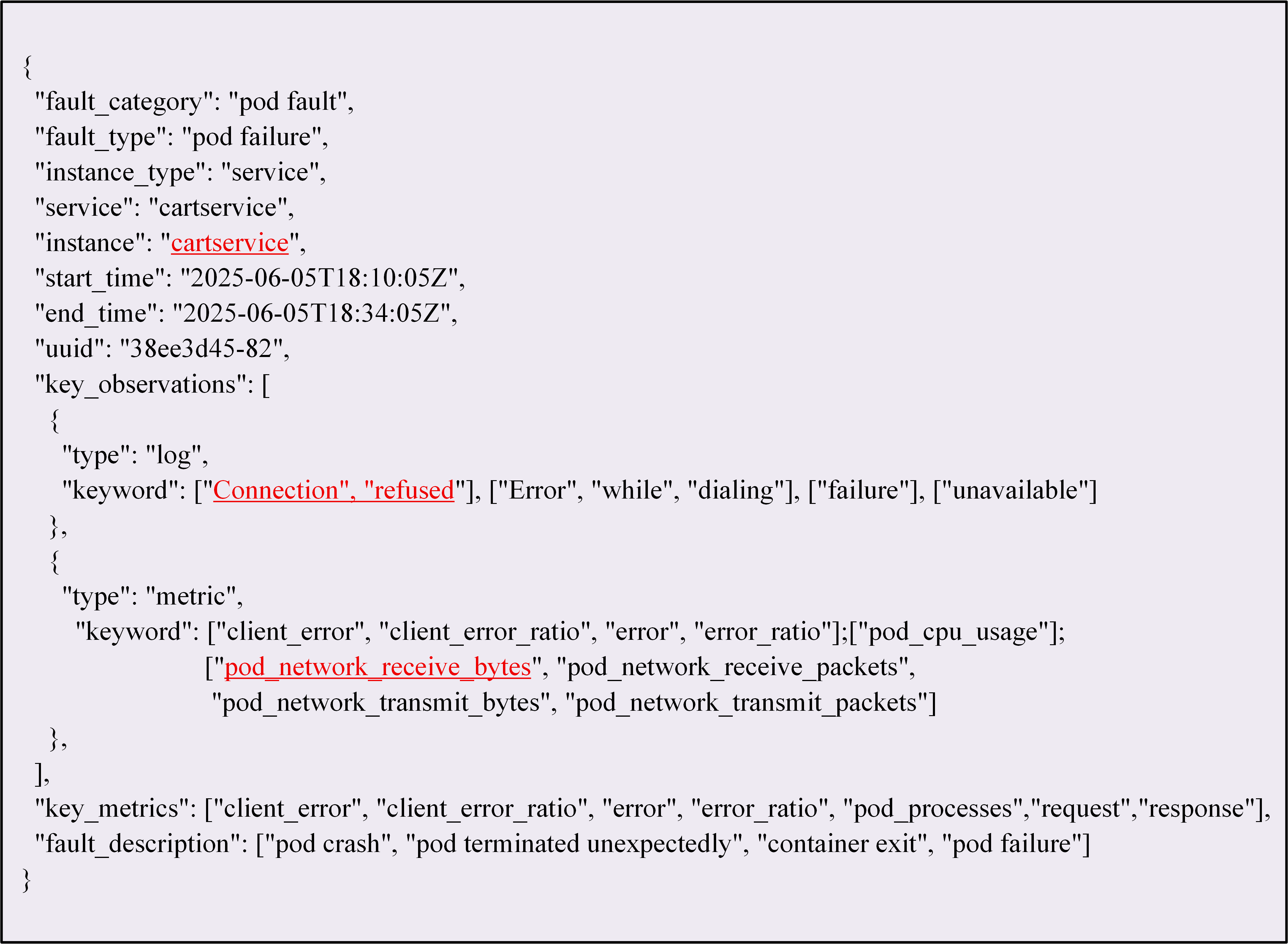}
    \caption{Bad case - ground truth range.}
    \label{fig21}
\end{figure*}

\begin{figure*}[!t]
    \centering
    \includegraphics[width=0.8\textwidth]{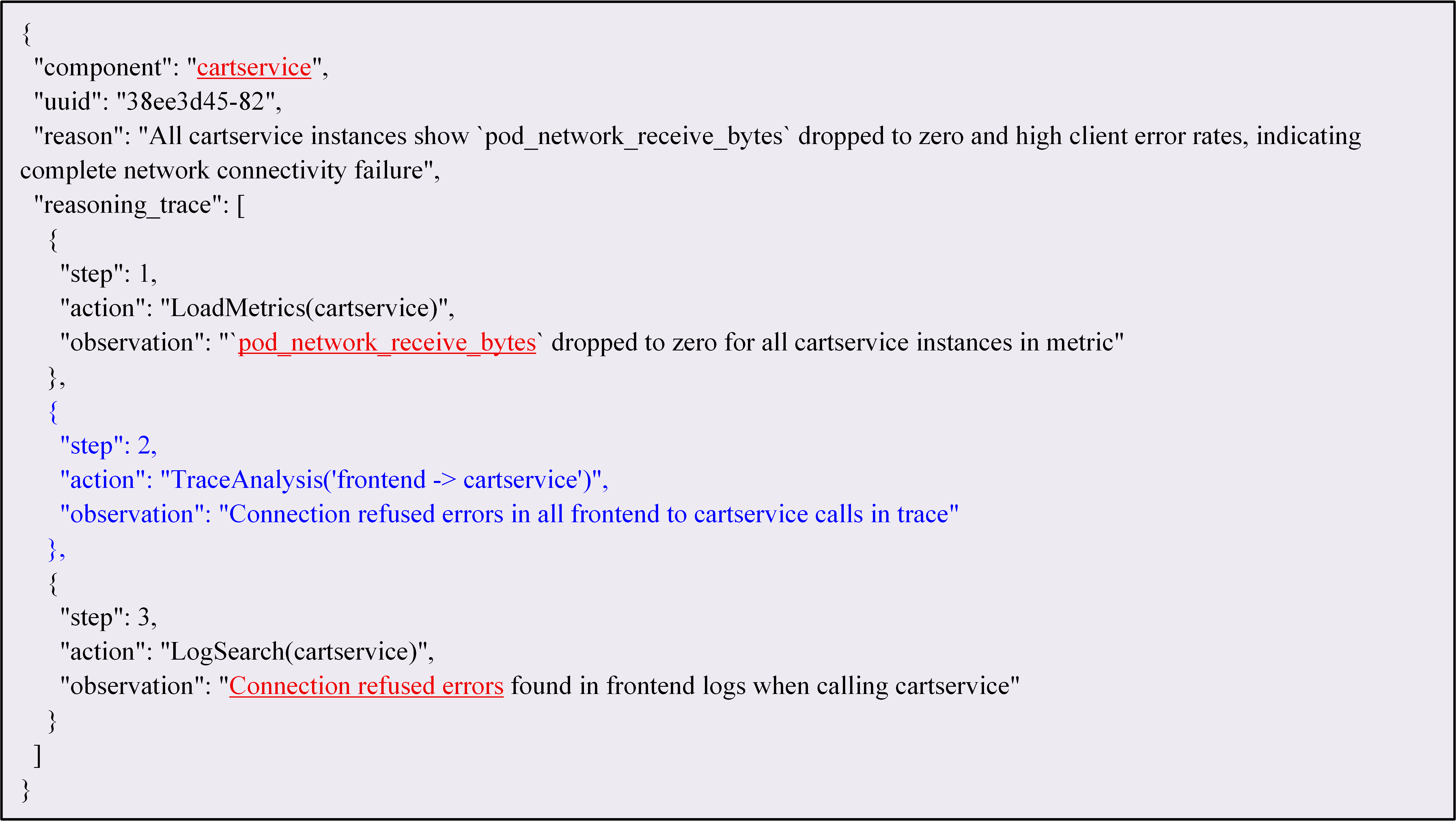}
    \caption{Bad case - actual output values.}
    \label{fig22}
\end{figure*}

Furthermore, we also explored typical bad cases to further identify areas for improvement in the proposed scheme. As shown in Figure \ref{fig21} and Figure \ref{fig22}, the large language model exhibited hallucination during the final root cause analysis (indicated in blue in the figures). The relevant trace call chains were not actually input into the large language model for root cause analysis, leading to an incorrect reasoning chain. This may result in deviations in the final fault root cause judgment and affect the stability of the scheme. In subsequent work, optimizations may be implemented by adding retrieval-augmented generation, jury mechanisms, and other methods. Additionally, after the competition released the answers, the log-related module can generate more effective keyword filtering templates through approaches such as enhancing the large language model's semantic judgment using broader and more representative fault data, thereby improving log filtering capabilities.

\subsection{Ablation Study}

\begin{table}[htbp]
\centering
\caption{Ablation Study of Each Component.}
\begin{tabular}{cccc}
    \toprule
    Log & Trace & Metric & Score \\
    \midrule
    \textcolor{red}{\checkmark} & $\times$ & $\times$ & 23.59 \\
    \midrule
    $\times$ & \textcolor{red}{\checkmark} & $\times$ & 31.09 \\
    \midrule
    $\times$ & $\times$ & \textcolor{red}{\checkmark} & 42.78 \\
    \midrule
    \textcolor{red}{\checkmark} & \textcolor{red}{\checkmark} & $\times$ & 35.32 \\
    \midrule
    $\times$ & \textcolor{red}{\checkmark} & \textcolor{red}{\checkmark} & 48.58 \\
    \midrule
    \textcolor{red}{\checkmark} & $\times$ & \textcolor{red}{\checkmark} & 51.27 \\
    \midrule
    \textcolor{red}{\checkmark} & \textcolor{red}{\checkmark} & \textcolor{red}{\checkmark} & 50.71 \\
    \bottomrule
\end{tabular}
\label{table1}
\end{table}

To further investigate the collaborative mechanism among the components of the proposed scheme, we conducted systematic ablation experiments. By comparing the performance of different modal combinations, we revealed the unique value and complementary relationships of each module. As shown in the experimental results in Table \ref{table1}, in the single-modal experiments, the metric module performed the most prominently (42.78 points), mainly due to its two-stage LLM analysis strategy that effectively covers full-stack monitoring perspectives across the service, pod, and node levels; the trace module ranked second (31.09 points), demonstrating strong capabilities in call chain analysis and identifying dependencies between services; while the log module exhibited relatively weaker performance when used independently (23.59 points), primarily because the unstructured nature of log data limits the accuracy of its standalone analysis. The dual-modal combination experiments revealed significant synergistic effects: the log+metric combination achieved the best performance (51.27 points), with this strong synergy stemming from the fact that metrics provide quantitative localization of performance anomalies while logs offer rich semantic descriptions of faults—together forming a perfect complementarity from numerical anomalies to semantic explanations; the trace+metric combination ranked next (48.58 points) as they share a natural correlation in the performance monitoring dimension; and the log+trace combination showed relatively weaker performance (35.32 points) due to the lack of quantitative support from the metric module. The complete three-modal fusion system achieved 50.71 points, which, although slightly lower than the 51.27 points of the optimal dual-modal combination (log+metric), still significantly outperformed other combinations—indicating that the synergy between logs and metrics can cover most fault patterns, while the trace module retains unique value in scenarios involving complex call chain anomalies. Ultimately, the results of the ablation experiments fully validate the scientific rigor of the multimodal fusion architecture in this scheme, confirm the key role of the metric module as the system’s core analysis engine, demonstrate the optimal performance of the log+metric combination, and provide important empirical support for constructing an accurate and interpretable solution for microservice fault root cause localization based on large-model agents.

\section{Conclusion}

This solution constructs a large-model agent analysis framework for fault root cause localization in microservice systems, which consists of five core modules: data preprocessing, log fault extraction, call chain anomaly detection, metric phenomenon summarization, and multimodal root cause analysis. First, the data preprocessing module realizes structured parsing of fault input and timestamp standardization, laying a foundation for multi-source data fusion. The log module leverages the Drain algorithm and a multi-level filtering mechanism to extract error templates and screen valid structured log features. The call chain module combines the Isolation Forest model with status code checks to perform dual detection of duration and status anomalies. The metric module filters core business and infrastructure metrics, and uses large language models to conduct two-level phenomenon summarization, generating highly readable anomaly descriptions. Finally, the multimodal root cause analysis module fuses the outputs of log, trace, and metric data, and drives large language models to conduct comprehensive reasoning through well-designed prompts, outputting structured root cause analysis results that include faulty components, causes, and reasoning trace. Ablation experiments fully verify the complementary value of multimodal data and the effectiveness of the system architecture. This framework performs excellently in complex microservice fault scenarios, ultimately achieving a score of 50.71.

\bibliography{custom}

\end{document}